\documentclass[sigconf]{acmart}
\usepackage{graphicx}       

\usepackage[utf8]{inputenc} 
\usepackage[T1]{fontenc}    
\usepackage{hyperref}       
\usepackage{url}            
\usepackage{amsfonts}       
\usepackage[linesnumbered,ruled,vlined]{algorithm2e}
\usepackage{nicefrac}       
\usepackage{microtype}      
\usepackage{fancyhdr}       
\usepackage{float}          
\usepackage{subfigure}      
\usepackage{xcolor}
\usepackage{multirow}
\usepackage{balance}

\usepackage{subcaption}

\usepackage[commandnameprefix=ifneeded, final]{changes} 

\AtBeginDocument{%
  \providecommand\BibTeX{{%
    \normalfont B\kern-0.5em{\scshape i\kern-0.25em b}\kern-0.8em\TeX}}}

\copyrightyear{2024}
\acmYear{2024}
\setcopyright{acmlicensed}
\acmConference[ICSE '24]{2024 IEEE/ACM 46th International Conference on Software Engineering}{April 14--20, 2024}{Lisbon, Portugal}
\acmBooktitle{2024 IEEE/ACM 46th International Conference on Software Engineering (ICSE '24), April 14--20, 2024, Lisbon, Portugal} \acmDOI{10.1145/3597503.3639181} 
\acmISBN{979-8-4007-0217-4/24/04}

\acmSubmissionID{12013}

\newtheorem{definition}{Definition}
\newtheorem{example}{Example}

\begin{document}

\title[MAFT: Efficient Model-Agnostic Fairness Testing for Deep Neural Networks]{MAFT: Efficient Model-Agnostic Fairness Testing for Deep Neural Networks via Zero-Order Gradient Search}


\author{Zhaohui Wang}
\affiliation{
    \institution{East China Normal University}
    \country{}
}
\email{51215902016@stu.ecnu.edu.cn}

\author{Min Zhang}
\authornote{Corresponding author.}
\affiliation{
    \institution{East China Normal University}
    \institution{Shanghai Key Laboratory of Trustworthy Computing}
    \country{}
}
\email{mzhang@sei.ecnu.edu.cn}

\author{Jingran Yang}
\affiliation{
    \institution{East China Normal University}
    \country{}
}
\email{nancyyyyang@163.com}

\author{Bojie Shao}
\affiliation{
    \institution{East China Normal University}
    \country{}
}
\email{51215902099@stu.ecnu.edu.cn}

\author{Min Zhang}
\affiliation{
    \institution{East China Normal University}
    \country{}
}
\email{zhangmin@sei.ecnu.edu.cn}

\renewcommand{\shortauthors}{Zhaohui Wang and Min Zhang, et al.}

\begin{abstract}

Deep neural networks (DNNs) have shown powerful performance in various applications and are increasingly being used \deleted{as} \added{in} decision-making systems. However, concerns about fairness in DNNs \deleted{are} always persist. 
Some efficient white-box fairness testing methods  about individual fairness have been proposed. \deleted{Nevertheless} \added{Nevertheless}, the development of black-box methods has stagnated, and the performance of existing methods is far behind that of white-box methods.
In this paper, we propose a novel black-box individual fairness testing method called Model-Agnostic Fairness Testing (MAFT). By leveraging MAFT, practitioners can effectively identify and address discrimination in DL models, regardless of the specific algorithm or architecture employed. 
Our approach adopts lightweight procedures such as gradient estimation and attribute perturbation rather than \deleted{heavy weight} \added{non-trivial} procedures like symbol execution, rendering it  significantly more scalable and applicable than existing methods.
We demonstrate that MAFT achieves the same effectiveness as state-of-the-art white-box methods  whilst\deleted{significantly reducing the time cost and} improving the applicability to large-scale networks. 
Compared to existing black-box approaches, our approach \deleted{generates up to 30 times more individual discriminatory instances with only 6.45\% - 22.57\% time cost.} \added{demonstrates distinguished performance in discovering fairness violations w.r.t effectiveness ($\sim14.69\times$) and efficiency ($\sim32.58\times$).}

\end{abstract}





\maketitle

\section{Introduction}
Deep learning (DL) has become an indispensable tool in various domains, including healthcare, finance, weather prediction, \deleted{and} image recognition and so on  \cite{esteva2019guide,heaton2017deep,pak2017review,ren2021deep}.
However, as \deleted{these deep} \added{DL} models increasingly influence critical aspects of \deleted{people's} \added{human} lives, it has been found that they are vulnerable to slight perturbations  \cite{goodfellow2014explaining, kurakin2016adversarial, papernot2016limitations, szegedy2013intriguing}. 
In addition to robustness risks, Deep Neural Networks (DNNs) also face fairness threats. \deleted{Deep learning} \added{DL} fairness is a growing area of concern that aims to address \deleted{biases} \added{bias} and unfairness in \deleted{deep models} \added{DNNs}. \deleted{Nowadays, it is essential to analyze the fairness violation of their results before the full acceptance of DNNs.}

Discrimination in \deleted{deep models} \added{DNNs} may arise from multiple sources, such as biased training data, inadequate features, or algorithmic discrepancies \cite{pessach2022review}. 
\deleted{Fairness means non-discrimination, including but not limited to group fairness and individual fairness, which means there are no group or individual will face unfair treatment due to their inherent properties. We call these properties protected attributes or sensitive attributes, such as race, gender and age. 
A biased or unfair model can lead to discrimination and unjust treatment of groups or individuals, exacerbating existing societal inequalities. For instance, a medical diagnosis model influenced by a patient's ethnicity or gender, rather than their actual medical condition, may result in misdiagnoses or \deleted{inadequate} \added{unreasonable} treatment plans. 
Therefore, it is crucial to develop methods and techniques that ensure the fairness of \deleted{deep models} \added{DNNs} for providing equitable services.}
To detect and evaluate software bias, a growing number of studies have been targeted at group fairness and individual fairness. 
Individual fairness\cite{cynthia2012fairness, galhotra2017fairness}emphasizes treating similar individuals similarly, regardless of what a protected group they belong to. It evaluates the model on the level of individual instances\added{,} which is more fine-grained and can capture subtle biases that may be ignored by the former\added{\cite{galhotra2017fairness}}. 
Thus we focus on individual fairness to develop methods that can effectively identify and address biases in \deleted{deep models} \added{DNNs} at the level of individual instances by generating a large number of individual discriminatory instances. Those instances can be used to retrain models to ease the discrimination of \deleted{deep models} \added{DNNs}. There have been \deleted{multiple} \added{several} relevant attempts \cite{agarwal2018automated, galhotra2017fairness, huchard2018proceedings} on the problem.

In the traditional field of machine learning, existing black-box fairness testing methods perform well.
Galhotra et al. proposed THEMIS which randomly samples each attribute in the neighborhood and identifies biased instances to measure the frequency of discrimination\cite{galhotra2017fairness}.
Udeshi et al. first proposed a two-stage global probabilistic search method called AEQUITAS to search discriminatory instances \cite{huchard2018proceedings}. 
Agawal et al. proposed a method called Symbolic Generation(SG), which combines two well-established techniques: local interpretability and symbolic execution \cite{aggarwal2019black}. 

However, these methods often shows low efficiency when they struggle to apply to \deleted{deep models} \added{DNNs}, given the architectural and algorithmic differences between machine learning and deep learning. The unique challenges posed by \deleted{deep models} \added{DNNs}, such as the complexity and non-linearity of their decision boundaries, demand novel efficient approaches to assess and mitigate biases. 

In this case, Zhang et al. proposed a\deleted{n} fairness testing approach \added{ADF} for \deleted{deep network models} \added{DNNs}\deleted{:  Adversarial Discrimination Finder (ADF)}. This approach still combines two-phase generation to search individual discriminatory instances based on input-specific probability distribution which depends on model internal information \cite{zhang2020white}.
Then, Zhang et al. confirmed that ADF is far from efficient, and further proposed \deleted{the Efficient Individual Discriminatory Instances Generator (EIDIG)} \added{EIDIG} to systematically generate instances that violate individual fairness. EIDIG inherits and further improves ADF to make it more effective and efficient\cite{zhang2021efficient}.
However, white-box approaches \deleted{like}\added{such as} ADF and EIDIG \deleted{cannot scale to large models and they} are \deleted{neither}\added{not} applicable \deleted{when models are agnostic} \added{in model-agnostic scenarios}. 


To this end, we propose a novel black-box individual fairness testing method called Model-Agnostic Fairness Testing (MAFT) which can be used in \deleted{deep model} \added{DNN} fairness testing without requiring access to their internal workings.
\deleted{We will explore bias in deep models and focus on the individual fairness metric, which evaluates the fairness of deep models at the level of individual instances.} MAFT inherits the workflow from EIDIG, so it is almost the same efficient as EIDIG. By converting the use of a real gradient into an estimated gradient, MAFT removes EIDIG's dependency on model, making it a model-independent black-box method for fairness testing \deleted{in the deep learning domain} \added{of DL}.  Despite the growing \deleted{interest} \added{interests} in \deleted{deep learning} \added{DL} fairness, there are only a few black-box fairness methods available that work effectively on deep models, making MAFT a valuable addition to the field.


We have implemented our framework MAFT and compared it with \added{both advanced} \deleted{white box} \added{white-box} methods \deleted{ADF and EIDIG} \added{and black-box methods}.
\deleted{According to ADF, its results show that they can generate\deleted{s} much more individual discriminatory instances (25 times) using much less time (half to 1/7) than existing black-box methods such as AEQUITAS and SG\cite{zhang2020white}. Compared with ADF, the number of instances generated by EIDIG is increased by 19.11\% with a speed up of 121.49 \% \cite{zhang2021efficient}. Our experimental results show that \deleted{the number and speed of discrimination instances generated by MAFT} \added{the effectiveness and efficiency of discrimination generation of MAFT} are almost the same to EIDIG. Then we can compute to get that our approach generates many more individual discriminatory instances (nearly 30 times) in much less time (6.45\% to 22.57\%) than current black-box approaches.}
\added{The overal improvement of MAFT over ADF is 7.92\% and 70.77\% in effectiveness and seperately, consistent with state-of-the-art EIDIG, outperform black-box methods AEQUITAS and SG.
MAFT demonstrates a substantial improvement over AEQUITAS and SG, achieving an increase of 1369.42\% in effectiveness over AEQUITAS and a 3158.43\% enhancement in efficiency compared to SG. These results highlight MAFT's significant advancements in black-box fairness testing domain.}

Overall, we make the following main contributions:
\begin{itemize}
    \item We propose a model-agnostic approach MAFT to do fairness testing for different models without knowing the inner information of model\added{s}. This versatility allows for broader \deleted{applicability} \added{applications} across different DNN systems.
    \footnote{https://github.com/wangzh1998/MAFT}.
    \item We evaluate MAFT against \deleted{the} white\added{-}box methods ADF and EIDIG\added{, along with black-box methods AEQUITAS and SG} with \deleted{10} \added{14} benchmarks on \deleted{three} \added{seven} real-word dataset\added{s}\deleted{, including four combinatorial benchmarks}. The experimental results show that MAFT is almost \deleted{as} \added{the} same as the state-of-the-art white-box method EIDIG in performance and \deleted{thus} far outperforms current black-box methods.
\end{itemize}

\deleted{The rest of the paper is organized as follows. 
In Section \ref{sec:background}, we briefly present the necessary background. 
In Section \ref{sec:methodology}, we describe the methodology of MAFT in detail. 
In Section \ref{sec:experiments}, we discuss our experimental setup \deleted{and} results \added{and limitations}. 
In Section \ref{sec:related work}, we review related work. 
Finally, in Section \ref{sec:conclusion}, we conclude our work. }

\section{Background}
\label{sec:background}

\subsection{Deep Neural Networks}

A deep neural network \deleted{$D$ architecture} usually contains an input layer for receiving input data, multiple hidden layers for learning, and one output layer for formatting the outputs. \deleted{An simple example is shown in Fig \ref{fig:DNN-A}.}

Usually, we can view a trained \deleted{deep neural network $\mathcal{D}$} \added{DNN} as a composite function $F(x)$ and compute its gradient using the chain rule implemented as backpropagation \deleted{in deep model} easily. The Jacobian matrix of $F(x)$ w.r.t a specific $x$ can be expressed as \deleted{in}~\cite{zhang2021efficient}:
\begin{equation}
    J_{F(x)} 
    = \frac{\partial F(x)}{\partial x}
    = \left[ \frac{{\partial F_m(x)}}{{\partial x_n}} \right]_{\added{n\times m}}
\end{equation}
where the $m$-th line is the gradient vector of the $m$-th output element respect to input data $x$.

In many cases, the information within hidden layers of a neural network remains unknown except input data and output confidence\deleted{as shown in Fig \ref{fig:DNN-B}}, making it \deleted{challenging to apply acclaimed white-box methods in fairness field} \added{intractable for white-box methods} such as ADF and EIDIG. 
However, our proposed approach MAFT \deleted{circumvents the need for leveraging hidden layer information. 
Instead, it} solely relies on knowledge of input data and output probabilities to find individual discriminatory instances, yielding excellent performance.

\subsection{Individual Discrimination}

We denote $X$ as a dataset and its attributes set $ A = \{A_1, A_2, ... ,A_n\}$. The input domain is denoted as $I$ and $I = I_1 \times I_2 \times ... \times I_n$ if each attribute $A_i$ has a value domain $I_i$. Then we use P to denote protected attributes of dataset $X$ like age, race, or gener, and obviously that $P \subset A$ and use $NP$ or $A \setminus P$ to denote non-protected attributes. As for a DNN model trained on $X$ that may \deleted{includes} \added{include} discrimination, we use $D(x)$ to denote its output label on $x$. 

\begin{definition}
\label{def:individual fairness}
Let $x = {[x_1, x_2, \ldots, x_n]}$, where $x_i$ is the value of attribute $A_i$, is an arbitrary instance in $I$. We say that $x$ is an individual discriminatory instance of a model $D$ if there exists another instance $x' \in I$ which satisfies the following conditions:
\begin{enumerate}
\item $\forall q \in NP$, $x_q = x'_q$;
\item $\exists p \in P$, \deleted{such that} $x_p \neq x'_p$;
\item $D(x) \neq D(x')$.
\end{enumerate}
Further, $(x, x')$ is called an individual discriminatory instance pair. Both $x$ and $x'$ are individual discriminatory instances.
\end{definition}

\begin{example}
    Let's consider a dataset about person information with 10 attributes, where the gender of a person is chosen as the protected attribute. We have the following pair $(x, x')$ from the dataset as an example:
    \begin{displaymath}
       \begin{split}
            x: [0, {\color{red}1}, 30, 1, 2, 0, 0, 1, 1, 0] \\
            x': [0, {\color{red}0}, 30, 1, 2, 0, 0, 1, 1, 0]
       \end{split}
    \end{displaymath}
    We highlight the gender attribute \deleted{whose index in the attribute vector is 1} in red for clarity. Except gender, $x$ and $x'$ have the same feature. If the decision-making \deleted{software} system \deleted{will provide} \added{provides} different prediction \deleted{label} \added{labels} for them, it would be thought as making decisions based solely on gender and ignoring any other attributes. As a result, $x$ would be considered as an individual discriminatory instance with respect to gender. \deleted{because it violates fairness property. Also, $(x, x')$ would be considered as an individual discriminatory instance pair.}
\end{example}

\subsection{Adversarial Attack}
\label{subsec:adversarial attack}
Recently, researchers have discovered that DNNs are \deleted{prone} \added{vulnerable} to robustness issues. Even state-of-the-art models can be easily deceived when attackers \deleted{generate attack inputs} \added{tamper the original input with minor distortion} that are unrecognizable to humans. In light of this, various adversarial attack methods have been proposed to enhance model robustness. These methods have also been found useful in other domains for generating adversarial samples to meet specific requirements, such as fairness testing\added{\cite{zhang2020white, zhang2021efficient}}.

Gradient-based adversarial attacks leverage gradients to minimize changes to the input while maximizing changes to the output of the sample. 
In \cite{goodfellow2014explaining}, FGSM(Fast Gradient Sign Method) is first proposed as a one-step attack that perturbs the input data by adding small perturbation in the direction of the gradient of the loss function with respect to input according to the following Equation:. 
\begin{equation}
    x^{adv} = x + \epsilon \cdot \mathbf{sign} (\nabla_x L(x,y))
\end{equation}
where $L$ is loss function of $D$, $y$ is predicted class given by $D(x)$ and $\nabla_x L(x,y)$ is gradient of loss function $L$ on $x$ w.r.t label $y$.
The perturbation is scaled by a small value noted by $\epsilon$ which controls the magnitude of the perturbation and real direction is only determined by the sign of gradient. FGSM is known for its simplicity and effectiveness in generating adversarial examples. 
\deleted{
Then BIM(Basic Iterative Method) was proposed in \cite{kurakin2016adversarial} as an extension of FGSM, and it is also known as the Iterative Fast Gradient Sign Method (IFGSM). In each iteration, BIM applies small perturbation to input data, and the process is repeated until a desired level of perturbation is reached. As a result, BIM can generate stronger adversarial examples. In \cite{dong2018boosting}, the author integrated momentum into BIM to generate more adversarial examples. In \cite{papernot2016limitations}, the author proposed a novel method called JSMA(Jacobian-based Saliency Map Attack). It identifies the most salient pixels in the input by computing salience map, which represents the sensitivity of the model's output to changes in each input pixel. The construction of salience map is given by:
\begin{equation}
\deleted{
    S_{xy}[i]= 
    \begin{cases} 
    0 \ if \ sign(J_{it}(x)) = sign(\sum_{j \neq t}J_{ij}(x)) \\
    J_{it}(x)*|\sum_{j \neq t} J_{ij}(x)| \ otherwise
    \end{cases}
    }
\end{equation}
}
\deleted{
where $J$ is the Jacobian matrix of the output y w.r.t the input x, and $t$ is the target class for the attack. 
JSMA aims to create adversarial examples that are less perceptible to humans while still causing misclassification by the model.
Different from FGSM and BIM, JSMA adopts the gradient of the model output instead of the loss function, which omits the backpropagation through loss function at each iteration. Inspired by this, a precise mapping between inputs and outputs of DNNs can be established with less time.}
\added{
Some iterative versions extending FGSM \cite{kurakin2016adversarial, dong2018boosting} were later proposed. Different from these methods, JSMA(Jacobian-based Saliency Map Attack) \cite{papernot2016limitations} adopts the gradient of the model output instead of the loss function, which omits the backpropagation through loss function at each iteration. Inspired by this, a precise mapping between inputs and outputs of DNNs can be established with less time.
}

In \cite{guo2019simple}, the author proposed a simple black-box adversarial \deleted{attacks} \added{attack}. By randomly sampling a vector from a predefined orthonomal basis and then either \deleted{add} \added{adding} or \deleted{subtract} \added{subtracting} it to the target image, the \deleted{DNNs} \added{DNN} output could be changed. Similar to the setup for training alternative models, the author of \cite{chen2017zoo} proposed a novel black-box attack that also only has access to the input (image) and output (confidence level) of the target DNN. Inspired by this, an efficient vectored gradient estimation method \deleted{be} \added{is} proposed in our work to guide the generation of discriminatory individual \deleted{instance} \added{instances}.

\subsection{Fairness Testing Problem Definition}

Before introducing a fairness testing problem, we would like to give a formal definition of perturbation \added{on non-sensitive attributes}.

\begin{definition}
    We use $x$ to denote the seed \deleted{that be perturbated} \added{input} and $p(x)$ to denote the \deleted{instances} \added{instance} generated from x by perturbation.
    We define the perturbation function as follow:
    \begin{equation*}
        p: I \times (A \setminus P) \times \Gamma \rightarrow I
    \end{equation*}
    where \deleted{$I$ is the input space of $x$, $(A \setminus P)$ means the perturbation is only add on non-sensitive attributes and} $\Gamma$ is the set of possible directions for perturbation, e.g., $\Gamma$ is defined as $\{-1,1\}$ for a single discrete attribute.
\end{definition}

A system \deleted{would like to} \added{tends to} make discriminatory decision when the system \deleted{has} \added{encodes} individual discrimination. Based on this, we define fairness testing problem as follow:

\begin{definition}
    Given a dataset $X$ and a DNN model $D$, we attempt to generate as many diverse individual discriminatory instances which violates fairness principle in $D$ (they can be used to mitigate discrimination) by perturbing the seed inputs in the dataset.
\end{definition}

\section{Methodology}
\label{sec:methodology}

\begin{figure*}
    \centering 
    \includegraphics[scale=0.4]{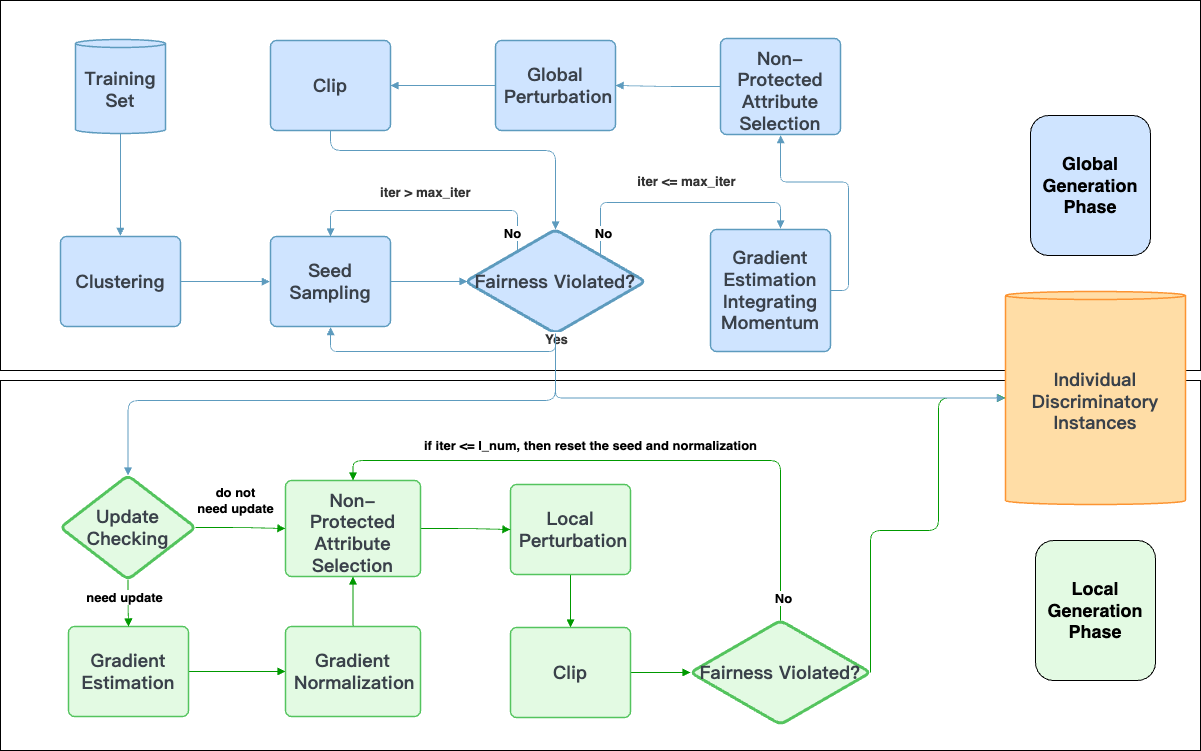}
    \caption{MAFT workflow to generate individual discriminatory instances inherited from EIDIG. }
    \label{Fig:workflow}
\end{figure*}


In this section, we present our algorithm called Model Agnostic Fairness Testing (MAFT) for generating individual discriminatory instances. We first describe how to \deleted{get the estimated gradient} \added{estimate gradient} from the black-box model for further gradient-based two-phase algorithm. 
Then we introduce total workflow of MAFT \deleted{which is inherited from EIDIG} shown in Fig \ref{Fig:workflow}.

\subsection{Zero-Order Gradient}
To leverage the gradient of any general black-box model $f$ for fairness testing, we employ the asymmetric difference quotient to estimate the gradient. 



We first approximate the derivative of function $f$ with a small \deleted{finite}\added{constant} \deleted{$\epsilon$} \added{$h$}:
\begin{equation}
    f'(x) = \lim_{ h \rightarrow 0} \frac{f(x+h)-f(x)}{h} \approx \frac{f(x+h)-f(x)}{h}
\end{equation}

Furthermore, we can extend this expression to \deleted{deep learning models $f$ which inputs has} \added{the predicted probability vector f of DNNs whose inputs consist of} multi-dimensional features\cite{goodfellow2016deep}:
\begin{equation}
       g_i := {\frac{\partial f(x)}{\partial x_{i}}} \approx \frac{f(x+h e_i) - f(x)}{h}, \label{equation:g_i} 
\end{equation}
where $g_i$ is the gradient $\frac{\partial f(x)}{\partial x_i}$\deleted{, $h$ is a small constant equal to $\epsilon$,} and $e_i$ is a standard basis vector wit h only the $i$-th component set to 1.

So we can get \deleted{estimation} \added{estimated} gradient $\Tilde{g_i}$:
\begin{equation}
{\Tilde{g_i}} :=  \frac{f(x+h e_i) - f(x)}{h} \label{equation:g_i}
\end{equation}

\begin{example}
    As an illustration of \deleted{the use of} the asymmetric difference formula: $f ' (x) \approx (f (x + h) - f (x))/h $, we consider a simple one-dimensional function $f(x) = x^2$ and choose the perturbation size $h = 0.001$ to compute the estimated gradient of $f$ at $x=2$.
    By substituting $x=2$ into the difference formula, we can get the estimated gradient value step by step:
    $f'(2) \approx (f (2 + 0.001) - f (2))/0.001 = ((2 + 0.001) ^ 2 - 2 ^ 2)/0.001 = 4.001 $.
    We can see that the estimate is very close to the real gradient $4$ of \deleted{this function} $f$ at $x=2$.
\end{example}

\subsubsection{Estimation Error Analysis}

It is important to note that the estimation error is of order $O(h)$. We can expand $f(x+h e_i)$ using the first-order Taylor series around $x$:
\begin{equation}
    f(x+h e_i) = f(x) + f'(x) \cdot h e_i + O(h^2)
\end{equation}
where $f'(x)$ is the first derivative of $f(x)$, and $O(h^2)$ represents all higher-order terms that are of order $h^2$ or higher.
Now, substituting this expansion into the formula of $\Tilde{g_i}$:
\begin{equation}
\Tilde{g_i} = \frac{f(x) + f'(x) \cdot h e_i + O(h^2) - f(x)}{h} = f'(x) \cdot e_i + O(h)
\end{equation}
The estimation error on dimension $i$ is given by the difference between the true gradient and the estimated gradient:
\begin{equation}
\text{Error} 
    = f'(x) \cdot e_i - \Tilde{g_i} 
    = f'(x) \cdot e_i - (f'(x) \cdot e_i + O(h))
    \added{= O(h)}
\end{equation}
\deleted{Thus we have:}
\deleted{
}
We can see that the estimation error of the asymmetric zero-order gradient on dimension $i$ is proportional to $h$. This means that the error decreases linearly with the step size $h$.

However, the perturbation size $h$ must be chosen carefully to ensure that the perturbation is not rounded down too much by finite-precision numerical computations in practical phases.
Despite concerns about numerical precision, obtaining an accurate gradient estimate is often unnecessary for successful searching. For instance, the \deleted{Fast Gradient Sign Method (}FGSM\deleted{)}\added{\cite{goodfellow2014explaining}} only requires the sign of the gradient, not its exact value, to discover adversarial examples. Thus, even if our initial approximations may lack accuracy, they can \deleted{still achieve remarkably} \added{achieve} high success rates, as demonstrated by our experiments in Section \ref{sec:experiments}.

\subsubsection{Original Zero-Order Gradient Algorithm}

\begin{algorithm}
    \caption{Original Zero-Order Gradient}
    \label{alg:zoo_non_vec}
    \SetNlSty{}{}{:}
    \DontPrintSemicolon
    \SetAlgoNlRelativeSize{0}
    \KwData{$x$, \text{model}, $perturbation\_size$}
    \KwResult{Estimated Gradient}
    $h \gets perturbation\_size$\;
    $n \gets \text{length}(x)$\;
    $y\_pred \gets \text{model}(x)$\;
    $gradient \gets \text{empty array of size} \ n$\;
    \For{$i$ \textbf{in range} $n$}{
        $x\_perturbed \gets \text{copy}(x)$\;
        $x\_perturbed[i] \mathrel{+}= h$ \;
        $y\_perturbed \gets \text{model}(x\_perturbed)$ \;
        $\textit{gradient}[i] \gets (y\_perturbed - y\_pred) / h$ \;
    }
    \If{$\text{model}(x) > 0.5$}{
        \textbf{return} $\textit{gradient}$\;
    }
    \Else{
        \textbf{return} $-\textit{gradient}$\;
    }
\end{algorithm}

Based on zero-order gradients, we propose a naive original Non-Vectored Zero-Order Gradient Algorithm ~\ref{alg:zoo_non_vec}. The algorithm aims to estimate the gradient of a given black-box model $f$ at a given input instance $x$. It perturbs each \deleted{property} \added{attribute} of $x$ with a specified perturbation step size and then obtains model output corresponding to the perturbed instance (lines 7,8). The finite difference is then used to calculate the gradient in one dimension (line 9). It proceeds iteratively until gradient values are obtained in all dimensions (lines 5-9).




In this version, gradient estimation for each attribute is calculated in \deleted{a} \added{an} explicit loop which has obvious drawbacks. 
Its main disadvantage is computational inefficiency, especially for large models and high-dimensional data, where the computational time required increases linearly with the number of input attributes. 
If the input instance has $n$ attributes, \deleted{it takes $n$ loops \deleted{to forward propagation} \added{ for forward propagation of the original input and its perturbed variant}and \deleted{make} \added{makes} differential calculations to get the complete gradient resulting in increased computational overhead.}\added{
the process involves one forward propagation for the original input, followed by $n$ loops to perform forward propagation for each of the $n$ perturbed variants. Additionally, differential calculations are performed separately for each variant to obtain the complete gradient, resulting in increased computational overhead. This results in a total of 
$n+1$ forward propagation steps and $n$ differential calculations.
} 
Therefore, it is \deleted{very} \added{too} slow for real-world applications, especially when dealing with high-dimensional data.

\begin{example}
This time, suppose we have a two-dimensional function: \deleted{$f(x1, x2) = {x1}^2 + {x2}^2$} \added{$f(x_1, x_2) = {x_1}^2 + {x_2}^2$} and we still use $h=0.001$.
If we have an input instance \deleted{$x = [x1, x2] = [2, 3]$} \added{$x = [x_1, x_2] = [2, 3]$}which has \deleted{$2$} \added{two} attributes, we should compute the asymmetric finite difference for each dimension to get the complete estimated gradient: First, we compute \deleted{$g_1 \approx (f(x1 + h, x2) - f(x1, x2)) / h$} \added{ $g_1 \approx (f(x_1 + h, x_2) - f(x_1, x_2)) / h$} and then \deleted{$ g_2 \approx (f(x1, x2 + h) - f(x1, x2)) / h$} \added{$ g_2 \approx (f(x_1, x_2 + h) - f(x_1, x_2)) / h$}. Thus, the zero-order gradient for $x = [2, 3]$ is $[4.001, 6.001]$. 
\end{example}

In this example, we only need to compute twice to get the estimated gradient since the function $f$ has only two attributes ($x1$ and $x2$). 
Note that \added{it consumes much time} to get the value of $f(x1+h, x2)$ or $f(x1, x2 + h)$ which is a complete forward propagation in deep learning\deleted{and it consumes a lot of time}. 
However, in most scenarios, the number of attributes can be larger, often exceeding ten or even more, which means the entire model must \deleted{be propagated} \added{propagate an input} ten or more times to estimate the gradient.
The \deleted{time} efficiency of the original non-vectored algorithm strongly depends on the number of attributes, making it less desirable for high-dimensional input instances. We aim to overcome this limitation by proposing a more efficient vectored zero-order gradient algorithm.

\subsubsection{Vectored Zero-Order Gradient Algorithm}

\begin{algorithm}[t]
    \caption{Vectored Zero-Order Gradient}
    \label{alg:zoo}
    \SetNlSty{}{}{:}
    \DontPrintSemicolon
    \SetAlgoNlRelativeSize{0}
    \KwData{$x$, \text{model}, $perturbation\_size$}
    \KwResult{Estimated Gradient}
    $h \gets perturbation\_size $\;
    $n \gets \text{length}(x)$\;
    $E \gets h \cdot I_n$ \;
    $X \gets x + E$ \;
    $Y \gets \text{model}(X)$ \;
    $y\_pred \gets \text{model}(x)$ \;
    $gradient \gets (Y - y\_pred) / h$  \;
    $gradient \gets \text{reshape}(gradient)$ \;
    \If{$\text{model}(x) > 0.5$}{
        \textbf{return} $\textit{gradient}$\;
    }
    \Else{
        \textbf{return} $-\textit{gradient}$\;
    }
\end{algorithm}

To overcome the limitations of \added{the} original version algorithm and improve computational efficiency, we subsequently introduce \deleted{a} more efficient Vectored Zero-Order Gradient \added{in} Algorithm~\ref{alg:zoo}. 

This algorithm takes \deleted{a} instance $x$, a black-box model and a hyperparameter $perturbation\_size$ as its input arguments and \deleted{return} \added{returns} the estimated gradient \deleted{as result}. 
We first initialize $h$ with specified perturbation size and denote the feature numbers of the input instance $x$ as $n$ (lines 1,2).
If we want to complete the calculation of the entire estimated gradient with only a fixed \added{few} times of forward propagation, \deleted{we should to} \added{we'd better} obtain \added{the gradient of} model output confidence \deleted{level} on right class $p$, denoted as \deleted{variations} $\nabla{F_p(x)}$, \deleted{on the feature perturbations of different dimensions} \added{to a set of variant inputs with different features perturbed} at the same time. 
Note that every \deleted{variation} \added{element} of $\nabla{F_p(x)}$ on different \deleted{feature} \added{features} should be independent.
To achieve this goal, we first should \deleted{to} construct an input square matrix $X$, each row of \deleted{the square matrix} \added{which} is a \deleted{perturbed input instance} \added{variant of} vector $x$ which has been perturbed on the $i$-th feature (lines 3,4). 
For the model, each row in $X$ is an independent input instance, so the perturbation of each \deleted{other} \added{one} \deleted{is irrelevant to the effect of the result} \added{has no impact on the results of others}.
We first construct a diagonal matrix of $n \times n$, where the diagonal element value is $h$ (line 3). It \deleted{is then added to x} \added{then adds x row by row} to obtain targeted square matrix $X$ consisting of $n$ perturbed instances (line 4).
In addition to the necessary forward propagation to obtain the black-box model output confidence \deleted{level} corresponding to the original input $x$ (line 6), we only need to perform  forward propagation once to obtain the \deleted{new} confidence \deleted{level} corresponding to each perturbed instance without knowing any internal structure of this model(line 5).
Then we use the \deleted{difference} \added{differential} calculation to get a \deleted{new} column vector, where each component is the \deleted{difference value of} \added{derivative to} the corresponding perturbation and reshape it to get the zero-order gradient we need (lines 7,8).
Finally, we \deleted{decide to return a positive layer or a negative layer} \added{calibrate the sign of the estimated gradient} according to whether the confidence \deleted{level} is greater than 0.5 (lines 9-12).

The calculation in the above process does not need to rely on any internal information of the model or internal calculation process, it only needs to obtain the confidence level of the model output \deleted{$\nabla{F_p(x)}$} \added{$F_p(x)$} in the correct classification $p$. Therefore, the above process can be widely used in different \deleted{models} \added{scenarios} to estimate gradient.

On the other hand, by performing perturbations on entire vectors simultaneously, the vectored version streamlines the computation, leading to better performance\deleted{, especially in the context of deep learning applications with high-dimensional data and complex model}. It enables us to estimate the entire gradient with just one or a fixed number of forward propagation, regardless of the number of attributes\deleted{. This advancement}\added{, which} is crucial for practical applications, as it empowers us to efficiently conduct fairness testing and model analysis on large-scale datasets and complex models, making the fairness testing process much more feasible and time-\deleted{effective}\added{effi  cient}.
This efficiency improvement greatly enhances the effectiveness and practicality of our proposed algorithm for fairness testing in real-world deep scenarios.

\deleted{Suppose we have a $n$-dimensional($n >= 10 $) function which have vectored functions and inputs, the vectored implementation of the zero-order gradient approximation method completes the calculation of $n$ gradient components with only fixed times forward propagation. It allows for improving computational efficiency compared to non\-vectored alternatives(which should do forward propagation $n$ times), this remarkable improvement in computational efficiency allows us to reduce the computation time to approximately one-tenth compared to non-vectored alternatives.}In essence, the efficiency achieved by the vectored approach brings it close to the computational efficiency of directly utilizing the computational graph for backpropagation, as demonstrated in our experiments as shown in \ref{fig:fig2}. This efficiency enhancement is particularly significant for adversarial searching tasks involving black-box models, where computational speed is crucial for effective and timely exploration of potential adversarial examples.

\subsection{Two-Phase Generation}


\begin{figure}
\centering 
\subfigure[Global Generation Intuition]{
    \label{fig:global_intuition}
    \includegraphics[scale=0.1]{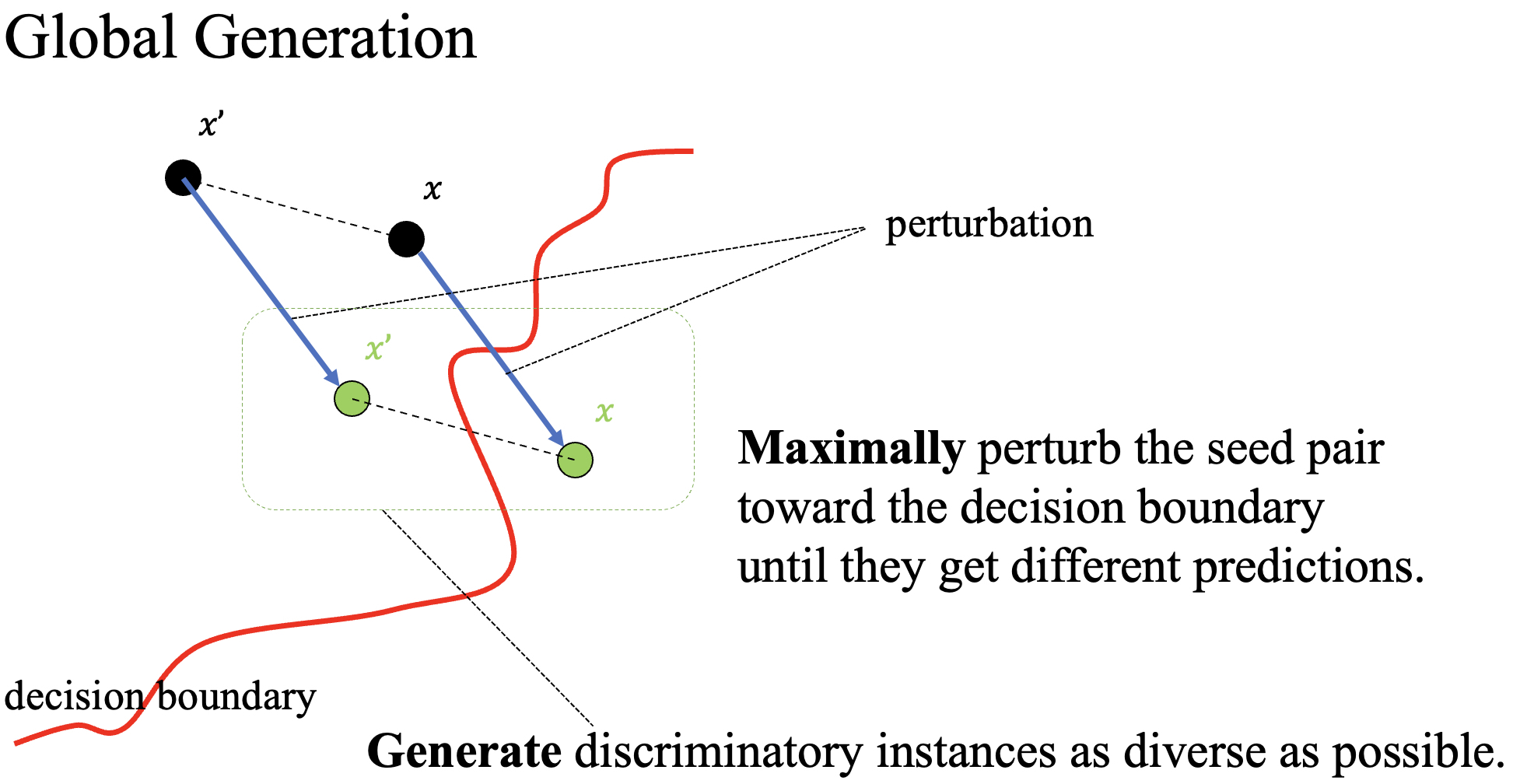}}
\subfigure[Local Generation Intuition]{
    \label{fig:local_intuition}
    \includegraphics[scale=0.1]{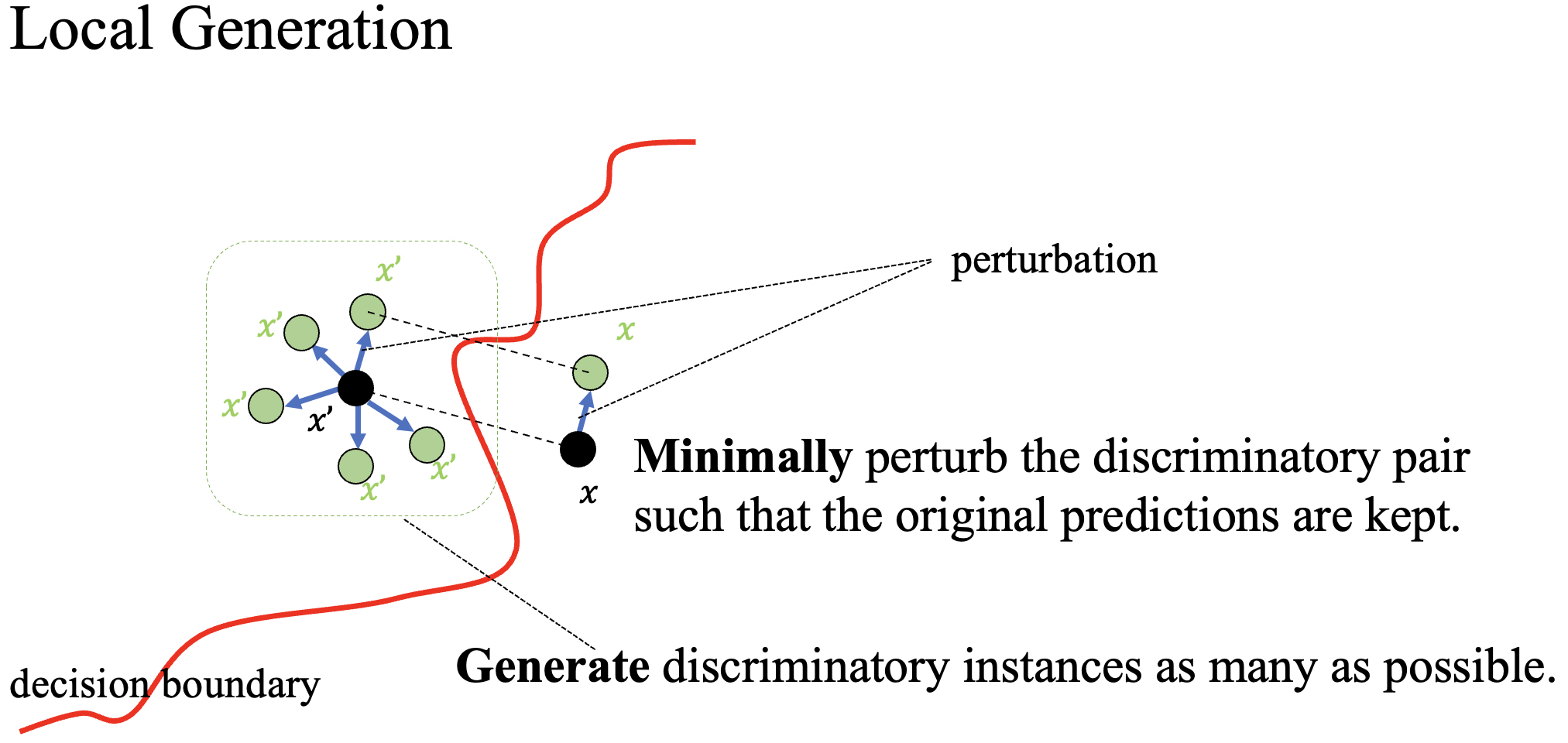}}
\vspace{-2mm}
\caption{\added{Two-Phase Generation Intuition}}
\label{Fig.intuition}
\end{figure}

In this section, we will introduce the workflow of MAFT, which consists of two sequential phases shown as Figure \ref{Fig:workflow} and discuss the improvement of MAFT \deleted{to transfer} \added{by transferring} it to the black-box fairness testing domain. 

\subsubsection{Global Generation}

\begin{algorithm}[b!]
    \caption{Global Generation}
    \label{alg:global_generation}
    \SetNlSty{}{}{:}
    \DontPrintSemicolon
    \SetAlgoNlRelativeSize{0}
    \KwData{Training set $X$, Attribute\deleted{s} set $A$, Protected attribute\deleted{s} set $P$, Input\deleted{s} domain $I$, \text{model}, $decay$, $max\_iter$, $global\_step$, $cluster\_num$, $global\_num$, $perturbation\_size$}
    \KwResult{Discriminatory Instances}
    $global\_id \gets \emptyset$ \;
     $clusters \gets \text{Clustering}(X, cluster\_num)$ \;
    \For{$i \gets 0$ to $global\_num-1$}{
         \text{Sample a seed $x$ from clusters in a round-robin fashion} \;
         $grad1 \gets \text{ZerosLike}(x)$ ;
         $grad2 \gets \text{ZerosLike}(x)$ \;
        \For{$iter \gets 0$ to $max\_iter - 1$}{
             $similar\_x \gets \{x' \in I \mid \exists a \in P, x'_a \neq x_a; $\;
             \hfill $\forall a \in (A \setminus P), x'_a = x_a\}$ \;
            \If{\text{IsDiscriminatory}$(x, similar\_x)$}{
                 $global\_id \gets global\_id \cup x$ \;
                 \textbf{break} \;
            }
             \deleted{$x' \gets \arg\max \{\left\|\text{ComputeGrad}(x) - \text{ComputeGrad}(x')\right\|_2 \mid x' \in similar\_x \}$ \;}
             \added{$x' \gets \arg\max \{\left\|\text{model}(x) - \text{model}(x')\right\|_2 \mid x' \in $\;
             \hspace{55pt} $similar\_x \}$ \;}            
             $grad1 \gets decay \cdot grad1 + \text{ComputeGrad}(x)$ \;
             $grad2 \gets decay \cdot grad2 + \text{ComputeGrad}(x')$ \;
             $direction \gets \text{ZerosLike}(x)$ \;
            \For{$a \in (A \setminus P)$}{
                \If{$sign(grad1[a]) = sign(grad2[a])$}{
                     $direction[a] \gets (-1) \cdot sign(grad1[a])$ \;
                }
            }
             $x \gets x + global\_step \cdot direction$ \;
             $x \gets \text{Clip}(x, I)$ \;
        }
    }
    \textbf{return} $global\_id$ \;
\end{algorithm}

Algorithm \ref{alg:global_generation} shows the procedure of global generation. We adopt the global generation phase to accelerate and diversify \deleted{the generated individual discriminatory instances} \added{discrimination generation}. \added{The intuition of it is shown in Fig \ref{fig:global_intuition}.} We first cluster the original input data set $X$ with clustering algorithms such as K-means \cite{lloyd1982least} with the goal of discovering diverse instances (line 2). We sample a seed \deleted{$X$} \added{$x$}  from the clusters in a round-robin fashion (line 4). For each selected seed, we perform $max\_iter$ iterations to find global discriminatory instances (lines 6-21).

According to Definition \ref{def:individual fairness}, we need to find an individual discriminatory instance pair to identify an individual discriminatory instance. 
So we get all instances that differ only in protected attributes from x as a set $similar\_x$(lines 7,8), whose size is the number of all possible combinations of the selected protected attributes in $I$ except $x$. 
Then we check whether $x$ violates individual fairness by \deleted{comparing}
\added{identifying} whether existing individual\added{s} in the $similar\_x$ have different label\added{s} with $x$. If so, $x$ can be added to $global\_id$ as a global discriminatory instance and the iteration stops(lines 9-11). 
Otherwise, we iteratively perturb $x$ until a new discriminatory instance is generated or $max\_iter$ is reached(lines 12-21).
\deleted{In other case, w}\added{W}e have to find a new discriminatory pair \deleted{$($x$, $x'$)$} \added{$(x,x')$}. 
To solve this problem, EIDIG chooses to traverse $similar\_x$ to select an instance $x'$ that has the biggest difference in \deleted{gradient} \added{model output}  with $x$(lines 12,13) because the intuition is that $x$ and $x'$ are more likely to be separated by the decision boundary of the model after perturbation if the Euclidean distance between \deleted{them} \added{their model predictions} is maximized.
Then, we perturb $x$ and $x'$ simultaneously on non-protected attributes to make \deleted{the former} \added{one of them} cross the decision boundary \deleted{on} \added{in} the opposite direction of the gradient by decreasing the model prediction confidence \deleted{level} on their original label (lines 16-20).
Momentum \cite{polyak1964some} is used as an optimization method for speeding up the procedure by accumulating local gradient and increasing efficiency, because it can help \deleted{stable} \added{stabilize} update and escape from \added{local} minimum or maximum\cite{duch1999optimization}(lines 14,15). 

EIDIG constructs a direct and precise mapping \deleted{$\nabla_{x}{F_{p}(x)}$} from \deleted{input feature $x$ and output variation $F_{p}(x)$} \added{input feature perturbation to output variation}\deleted{on original right label $p$}\added{,} which should be done by internal backpropagation of model \added{for gradient calculation}. However, we replace this computation of $\nabla_{x}{F_{p}(x)}$ with our estimated gradient shown as $ComputeGrad(x)$ to make this method independent of the model itself. 

At last, we clip the generated instance $x$ to keep it within the input domain $I$ (line 21).
Finally, this algorithm return\added{s} all generated discriminatory instances (i.e., $global\_id$) which will be used as the seed inputs for local generation phase.

\subsubsection{Local Generation}

\begin{algorithm}[t]
    \caption{Local Generation}
    \label{alg:local_generation}
    \SetNlSty{}{}{:}
    \DontPrintSemicolon
    \SetAlgoNlRelativeSize{0}
        \KwData{Attribute\deleted{s} set $A$, Protected attribute\deleted{s} set $P$, Input\deleted{s} domain $I$, $local\_num$, $global\_id$, \text{model}, $update\_interval$, $local\_step$}
        \KwResult{Discriminatory Instances}
            $local\_id \gets \emptyset$ \;
            \For{$x \in global\_id$}{
                 $suc\_iter \gets update\_interval$ \;
                \For{$iter \gets 0$ to $local\_num - 1$}{
                    \If{$suc\_iter \geq update\_interval$}{
                         $similar\_x \gets \{x' \in I \mid \exists a \in P, x'_a \neq x_a; $\;
                         \hspace{55pt} $\forall a \in (A \setminus P), x'_a = x_a\}$ \;
                         $x' \gets \text{FindPair}(x, similar\_x, model)$ \;
                         $grad1 \gets \text{ComputeGrad}(x)$ \;
                         $grad2 \gets \text{ComputeGrad}(x')$ \;
                         $probability \gets \text{Normalization}(grad1, grad2, P)$ \;
                         $suc\_iter \gets 0$ \;
                    }
                     $suc\_iter \gets suc\_iter + 1$ \;
                     $a \gets \text{RandomPick}(probability)$ \;
                     \deleted{$direction \gets [-1, 1]$ \;}
                     $direction \gets (-1, 1)$ \;
                     $s \gets \text{RandomPick}([0.5, 0.5])$ \;
                     $x[a] \gets x[a] + direction[s] \cdot local\_step$ \;
                     $x \gets \text{Clip}(x, I)$ \;
                     $similar\_x \gets \{x' \in I \mid \exists a \in P, x'_a \neq x_a; $\;
                        \hspace{55pt} $\forall a \in (A \setminus P), x'_a = x_a\}$ \;
                    \If{\text{IsDiscriminatory}$(x, similar\_x)$}{
                         $local\_id \gets local\_id \cup x$ \;
                        }
                    \Else{
                         $\text{Reset}(x)$ ;
                         $\text{Reset}(probability)$ \;
                         $suc\_iter \gets 0$ \;
                    }
                }
            }
            \textbf{return} $local\_id$ \;
\end{algorithm}

Algorithm \ref{alg:local_generation} shows the contents of local generation phase, which quickly generates as many discriminatory instances as possible around the seeds $global\_id$ generated by the global generation phase. \added{The intuition of it is shown as Fig \ref{fig:local_intuition}.}

Unlike the global phase, which maximizes \deleted{change} \added{output variation} to discover potential individual discriminatory instance pairs at each iteration, the local phase \deleted{minimizes} \added{focuses on changing} \added{model variation} \deleted{change} to maintain the original predictions from the model\added{.} \deleted{to get more similar discriminatory instances, motivated by robustness that similar inputs lead to similar or same outputs}\added{In doing so, we could get more similar discriminatory instances, which is motivated by DNN robustness that similar inputs lead to similar or the same outputs} \cite{du2020fairness, szegedy2013intriguing}. For this, we just need slight perturbation on the local phase. 

For each global seed from $global\_id$, there must be a similar instance $x'$ which has different label(lines 6-8). 
Then we also use gradient information to guide attribute selection and perturbation(lines 9-11, 14-17). 
To keep the prediction, we prefer to select an attribute that has less impact on the model\deleted{thus}\added{, and thus the} instance can keep the same label after perturbation. 
\deleted{Thus we}\added{We} calculate normalized probabilities of non-protected attributes as attribute contribution according to reciprocals of their gradients(line 11). 
Then we use this probability to choose an attribute to perturb on random direction $s$ (lines 14-17), because we tend to choose an attribute that has less impact on the model after perturbation. 
If $x$ becomes a new discriminatory instance after perturbation, we add it to $local\_id$ set \deleted{as final result}(lines 21,22). Otherwise, we reset $x$ and corresponding probability to next iteration.
EIDIG shows that the information guiding attribute selection and perturbation is likely to be highly correlated at each iteration due to small perturbation in the local generation phase, so it chooses to calculate the attribute contributions every few iterations(lines 5-12).

During this phase, EIDIG still use \added{$\nabla_{x}{F_{p}(x)}$ to establish} a direct and precise mapping \deleted{$\nabla_{x}{F_{p}(x)}$} from \deleted{input feature $x$ and output variation $F_{p}(x)$} \added{input perturbation to output variation} on original right label $p$ to compute \deleted{normalization} \added{normalized} probability. However, we replace this computation of $\nabla_{x}{F_{p}(x)}$ with our estimated gradient shown as $ComputeGrad(x)$ again to make this method independent of the model itself(lines 9,10). 

At this time, we have generated a large number of individual discriminatory instances, which can be used for retraining to remove bias from the original model.

\section{Implementation and Evaluation}
\label{sec:experiments}


In this section, we present experiments designed to evaluate the performance of MAFT and explore why estimated gradient is effective and efficient. The experiments can be structured into three primary research questions (RQs):
\begin{description}
    \item[RQ1:] How does the choice of hyperparameter affect the performance of MAFT?
    \item[RQ2:] Given the selected hyperparameter, how does \deleted{the} MAFT compare with \added{AEQUITAS,} ADF and EIDIG in terms of effectiveness and efficiency?
    \item[RQ3:] To what extent does the estimated gradient in MAFT match the real gradient in EIDIG in terms of effectiveness and efficiency?
\end{description}

\subsection{Experimental Setup}

\paragraph{Baselines}

\begin{table}
    \centering
        \caption{Comparing different methods.}
    \begin{tabular}{c|c|c|c|c|c|c}
        \toprule
        \deleted{Method Feature}& THE & AEQ & SG & ADF & EIDIG & MAFT \\
        \midrule
        Effective \& Efficient & $\times$ & $\times$ & $\times$ & $\checkmark$ & $\checkmark$ & $\checkmark$ \\
        Model-agnostic & $\checkmark$ & $\checkmark$ & $\checkmark$ & $\times$ & $\times$ & $\checkmark$ \\ 
        \bottomrule
    \end{tabular}
    \label{tab:methods_comparision}
\end{table}
According to the features of different methods that are shown in Table \ref{tab:methods_comparision} (THE and AEQ are the \deleted{short} \added{abbreviation} of THEMIS and AEQUITAS separately), we can know that model-agnostic methods such as \deleted{THEMIS, } AEQUITAS and SG are ineffective and inefficient \deleted{compared with white-box methods such as ADF and EIDIG and this has been proved in \cite{zhang2020white}}.
To answer these questions, we \added{s}elect to \deleted{only} utilize \added{AEQUITAS, SG,} ADF and EIDIG as baseline comparison techniques. 
\added{As THEMIS is shown to be significantly less effective \cite{aggarwal2019black} and thus is omitted.}
\added{Through comparison with black-box methods, we demonstrate the superior performance of MAFT. Additionally, by contrasting it with white-box methods, we highlight why it have distinct advantages and strengths.}
\deleted{As ADF and EIDIG present the optimal performance, we choose them as baseline ignoring other black-box methods. This decision allows us to focus our attention on comparing MAFT with the most advanced methods currently available.}

In addition to this, we have to highlight that MAFT differs from ADF and EIDIG by operating as a black-box approach, \deleted{therefore}\added{so} retraining is unnecessary to demonstrate the effectiveness of the discriminatory instances generated. If you wish to confirm the effectiveness of these instances in retraining to alleviate the model's bias, you are encouraged to \deleted{reference} \added{refer to} ADF and EIDIG.

\paragraph{Datasets and Models}

To evaluate MAFT, we select \deleted{three} \added{seven} benchmark datasets that have been \deleted{widely} used in previous studies\cite{aggarwal2019black, galhotra2017fairness, huchard2018proceedings, zhang2020white, zhang2021efficient}.
The details are as \deleted{following} \added{Table \ref{tab:datasets and model acc} ($\#Ins$ means number of instances, $\#Fea$ expresents the number of features and $Protected\ Attrs$ are protected features)}.
    \deleted{Census Income 
    This dataset contains 32561 training instances with 13 attributes with containing potential sensitive attributes like age, race and gender. }
    \deleted{German Credit 
    \deleted{This} is a small dataset with 600 data \deleted{ad} \added{and} 20 attributes. Both age and gender are sensitive attributes. }
    \deleted{Bank Marketing
    \deleted{The dataset} has more than 45000 instances with 16 attributes and the only protected attribute is age. }

\begin{table}
    \centering
        \caption{\added{Benchmark Datasets and Model Accuracy}}
    \begin{tabular}{c|c|c|c|c}
        \hline
        Dataset & \#Ins & \#Fea & Protected Attrs & Accuracy \\ \hline
        Bank Marketing \footnotemark[2] & 45211 & 16 & Age & 89.22\% \\ 
        Cencus Income \footnotemark[3] & 48422 & 12 & Age Race Gender & 84.32\% \\
        German Credit \footnotemark[4] & 1000 & 24 & Age Gender & 78.25\% \\
        Diabetes \footnotemark[5] & 768 & 8 & Age & 74.03\% \\
        Heart Heath \footnotemark[6] &  297 & 13 & Age Gender & 74.79\% \\
        MEPS15 \footnotemark[7] &  15830 & 137 & Age Race Gender & 84.55\% \\
        Students \footnotemark[8] & 1044 & 32 & Age Gender & 86.6\% \\ 
        \hline
    \end{tabular}
    \label{tab:datasets and model acc}
\end{table}

The predictive tasks based on these datasets center\deleted{ed} around determining whether an individual \deleted{met}\added{meets} certain conditions. 
Owing to the simplicity of these datasets, we employs \deleted{a} \added{the} six-layer fully connected network\added{s} that was \deleted{trained} \added{adopted} by EIDIG. \added{The details of the modes's accuracy are shown in Talbe \ref{tab:datasets and model acc}.}
Prior to the tasks' beginning, it was necessary to preprocess the data particularly in regard to the conversion of continuous attributes into categorical ones. For instance, we discrete age based on human lifecyle. 
For ease of denotation, each benchmark was denoted as "B-a", where "B" represents the initial uppercase letter of the dataset, and "a" refers to the initial lowercase letter of the sensitive attribute. \deleted{Benchmark of sensitive attribute\deleted{s} \deleted{subgroups}\added{combinations} would be \deleted{noted}\added{denoted} as "B-a\&c" where "c" represents another sensitive attribute.}

\paragraph{Configuration\added{s}}



Both ADF and EIDIG are configured according to the best performance setting reported in the respective papers. 
For \deleted{both} global \deleted{and local phase}, max iteration $max\_iter$ is set to 10 because less than 5 iterations need to be taken for most situations to find a\deleted{b}\added{n} individual discriminatory instance around the seed if \deleted{it is}\added{there} exists and both global step size \deleted{$s\_g$} \added{$global\_step$} and local step size \deleted{$l\_g$} \added{$local\_step$} are set to 1.0, i.e., the minimum step for the categorical attributes. During global phase, we set cluster count \deleted{$c\_num$} \added{$cluster\_num$} to 4 as ADF and EIDIG did to cluster the training set using K-Means \cite{lloyd1982least}. As EIDIG achieves best performance when past gradient information decays away to half its origin after each iteration, we set decay factor of momentum \deleted{$\eta$} \added{$decay$} to 0.5. During local phase, we set prior information life cycle $update\_interval$ to 5 to make a balance of prior information effectiveness and update \deleted{speed}\added{frequency}. As for perturbation size \deleted{$h$} \added{$perturbation\_size$}, we will discuss it in \ref{RQ1}.

\added{We re-implemented SG and AEQUITAS using TensorFlow 2 and make slight adjustments to make them having the same maximum search iterations with white-box methods under same parameters. We opt for a fixed random seed to generate initial input for AEQUITAS in input domain to keep stability. Other settings are also kept best.}

\deleted{All experiments were conducted on a personal computer equipped with an Apple M1 Pro chip, 16GB of unified memory, running macOS Ventura. }
\added{All experiments were run on a personal computer with 32 GB RAM, Intel i5-11400F 2.66GHz CPU and NVIDIA GTX 3060 GPU in Ubuntu22.04}

\footnotetext[2]{https://archive.ics.uci.edu/ml/datasets/bank+marketing}
\footnotetext[3]{https://www.kaggle.com/vivamoto/us-adult-income-update?select=census.csv}
\footnotetext[4]{https://dataverse.harvard.edu/dataset.xhtml?persistentId=doi:10.7910/DVN/Q8MAW8}
\footnotetext[5]{https://archive.ics.uci.edu/ml/datasets/diabetes}
\footnotetext[6]{https://archive.ics.uci.edu/ml/datasets/Heart+Disease}
\footnotetext[7]{https://github.com/HHS-AHRQ/MEPS}
\footnotetext[8]{https://archive.ics.uci.edu/ml/datasets/Student+Performance}

\subsection{Results and Discussion}

Notice that \added{besides AEQUITAS,} all of ADF, EIDIG and MAFT \deleted{comprise a global generation phase and a local generation phase} \added{share a similar gradient-based testing framework with two sequential phases}. To this end, we compare them phase by phase to answer these research questions.

\subsubsection{Hyperparameter Study: Perturbation Size (RQ1)}
\label{RQ1}


\begin{figure}
\centering 
\includegraphics[scale=0.28]{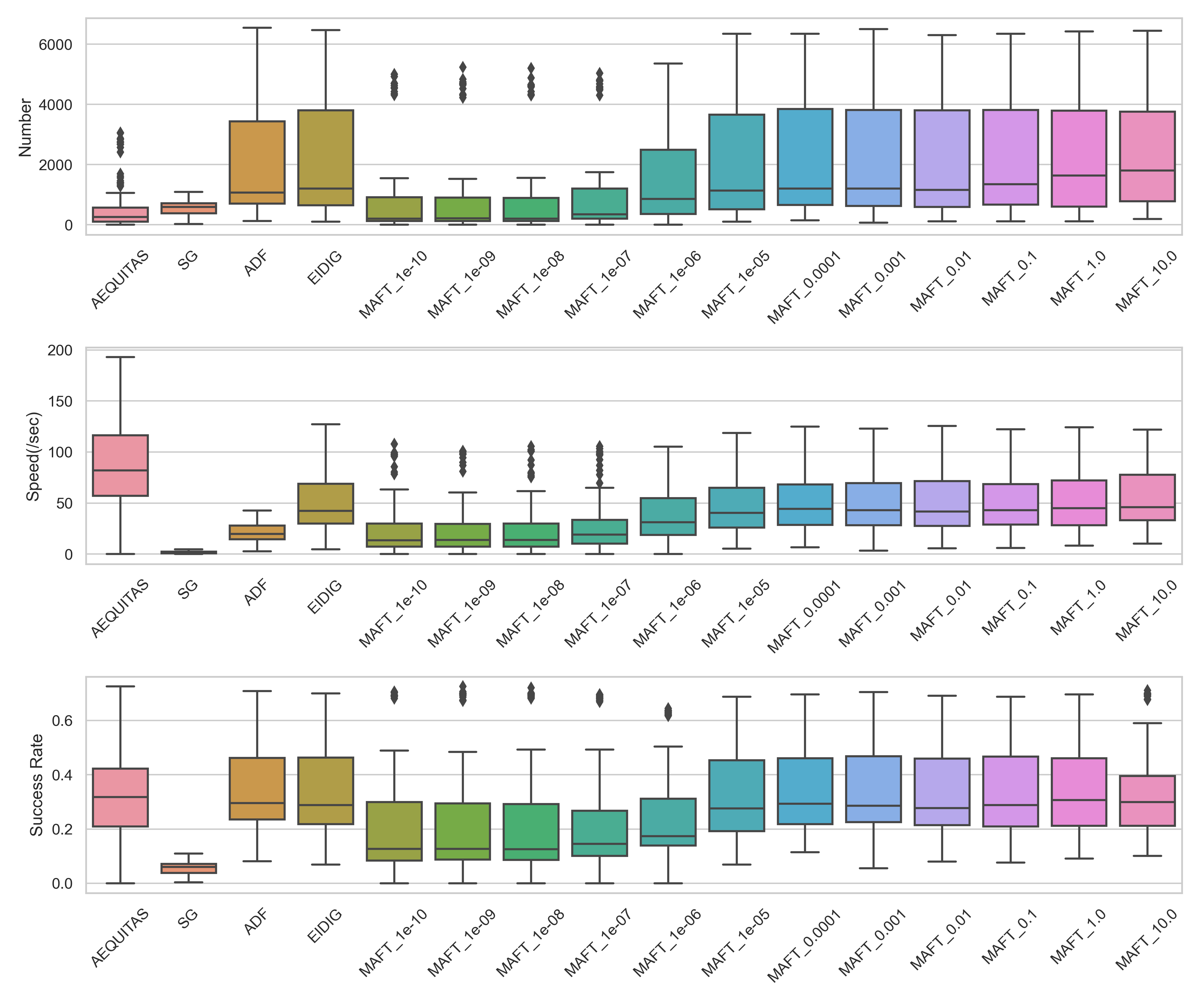}
\caption{\added{Hyperparameter: Perturbation Size Comparison}}
\label{Fig:hyper comparison}
\end{figure}


The perturbation size $h$ is a critical hyperparameter in MAFT \deleted{that controls the magnitude of perturbations applied to the input instances during exploration}. 
In this experiment, our goal is to provide valuable \deleted{insight to the affect}\added{insights for the effects} of perturbation size and determine the optimal perturbation size \deleted{hyperparameter} for the MAFT method across various benchmarks \added{in the settings of tabular data}. 
In these experiments, $g\_num$ and $l\_num$ were set to 100 and 100 separately (in the formal experiments, they were set to 1000 and 1000). 
We performed experiments on multiple benchmarks using five different methods: MAFT with various perturbation size values ranging from $1e-10$ to $10$, AEQUITAS, SG, ADF and EIDIG, each with their optimal parameters. 
Specifically, the methods included MAFT, varying across $12$ parameter configurations, thus yielding a total of $16$ methods of distinct settings. Each unique combination underwent rigorous evaluation through $10$ rounds. Subsequently, we aggregated the resultant data points for each method and plotted them with a boxplot shown in Fig \ref{Fig:hyper comparison}.

In Fig \ref{Fig:hyper comparison}, vertical axes in different \deleted{sub figures}\added{subfigures} represent\deleted{s} instance generation number, instance generation speed, and \deleted{iterations} \added{success rate} for generating discriminatory instances \added{in effective attempts} separately.
\added{Notice that higher values are better for all of the three metrics alone.}
The horizontal axis represents \added{AEQUITAS, SG,} ADF, EIDIG, and MAFT with different perturbation size values.
We get several observations from the results. 
\deleted{First, two curves of those three show similar trends (Notice that higher values are better for average instance number and average instance generation speed, while lower values are better for average iterations).}
\added{AEQUITAS adopts a global adaptive search strategy, which continuously reinforces past successful choices during search. In doing so, AEQUITAS is more inclined to exploit than to explore, and thus it stucks in duplicate search paths, resulting in finding much less fairness violations, even if it achieves good performance in terms of generation speed and success rate due to it heuristic nature.}
\added{SG is both ineffective and inefficient, because it heavily relys on the existing techniques about local explanation and symbolic execution, which are both time-consuming. Specifically, MAFT with $h=1$ generates 292.15\% more biased samples than SG at \deleted{3258.43\%} \added{3158.43\%} higher speed on average.}
EIDIG significantly outperforms ADF \deleted{in generation number and} \added{especially in} speed.
\added{Moreover, the success rate of EIDIG is slightly lower than ADF, since EIDIG doesn't utilizes completely accurate gradients in most iterations during local search, but EIDIG explores much more search space owing to its momentum integration during global search.}
For MAFT, when $h$ is less than or equal to \deleted{$1e-8$} \added{$1e-6$}, the results are even significantly inferior to ADF. 
However, the results improve\deleted{s} as $h$ increase\deleted{d}\added{s} within the range of $(1e-8, 1e-5]$ and keep increasing until \deleted{remains}\added{the metrics remain} relatively stable within the range of $(1e-5, 10]$ with the number and speed metrics \deleted{on par}\added{comparable} with EIDIG.
\added{We apply ANOVA\cite{girden1992anova} to verify that the experimental results of MAFT with $h$ in $(1e-5, 10]$ do not exhibit statistically significant differences.}
At $h=1$ \added{and $h=10$}, we get the highest average number and speed, but \deleted{these values significantly dropped when $h=10$} \added{the success rate at $h=10$ obviously drops}.
These results demonstrates that when $h$ is too small, the perturbation is so subtle that it fails to estimate an useful gradient. Conversely, when $h$ is too large, the estimated gradient deviates significantly\deleted{ negative affect on the results}. When $h$ is kept at an appropriate value, the experimental effects are relatively stable and efficient.
\deleted{As for the average iterations shown in final curve, lower values would generally be more desirable. The number of iterations may not be optimal at $h=1$ because \deleted{as the exploration space increases to produce more instances, }the average number of iterations must rise \added{as the exploration space increases to produce more instances}. For example, iterations of EIDIG are more than \deleted{Adf}\added{ADF} even though the former is more effective and efficient.}

\added{Overall, we recommand using the perturbation size of 1, and it is also the minimum granularity of the preprocessed attributes \added{of the subject tabular datasets} after discretization. We also encourage users to try to choose parameters that are more suitable for their specific datasets.}

\deleted{In conclusion, to maintain efficient experimental performance, we choose $h=1$ as the hyperparameter for MAFT. Although we could have chosen different hyperparameters for different benchmarks to achieve the best performance, we opted for simplicity and followed EIDIG's approach of using a global uniform parameter.}

\textbf{In a certain range, the influence of \deleted{hyperparameter}\added{perturbation magnitude} is stable. In practical applications, we can choose the optimal parameters on demand. For simplicity, we recommand using the perturbation size of 1 for preprocessed tabular datasets with discret attributes.}

\subsubsection{Comprehensive Results: Effectiveness and Efficiency (RQ2)}


\begin{figure}
\centering 
\includegraphics[scale=0.18]{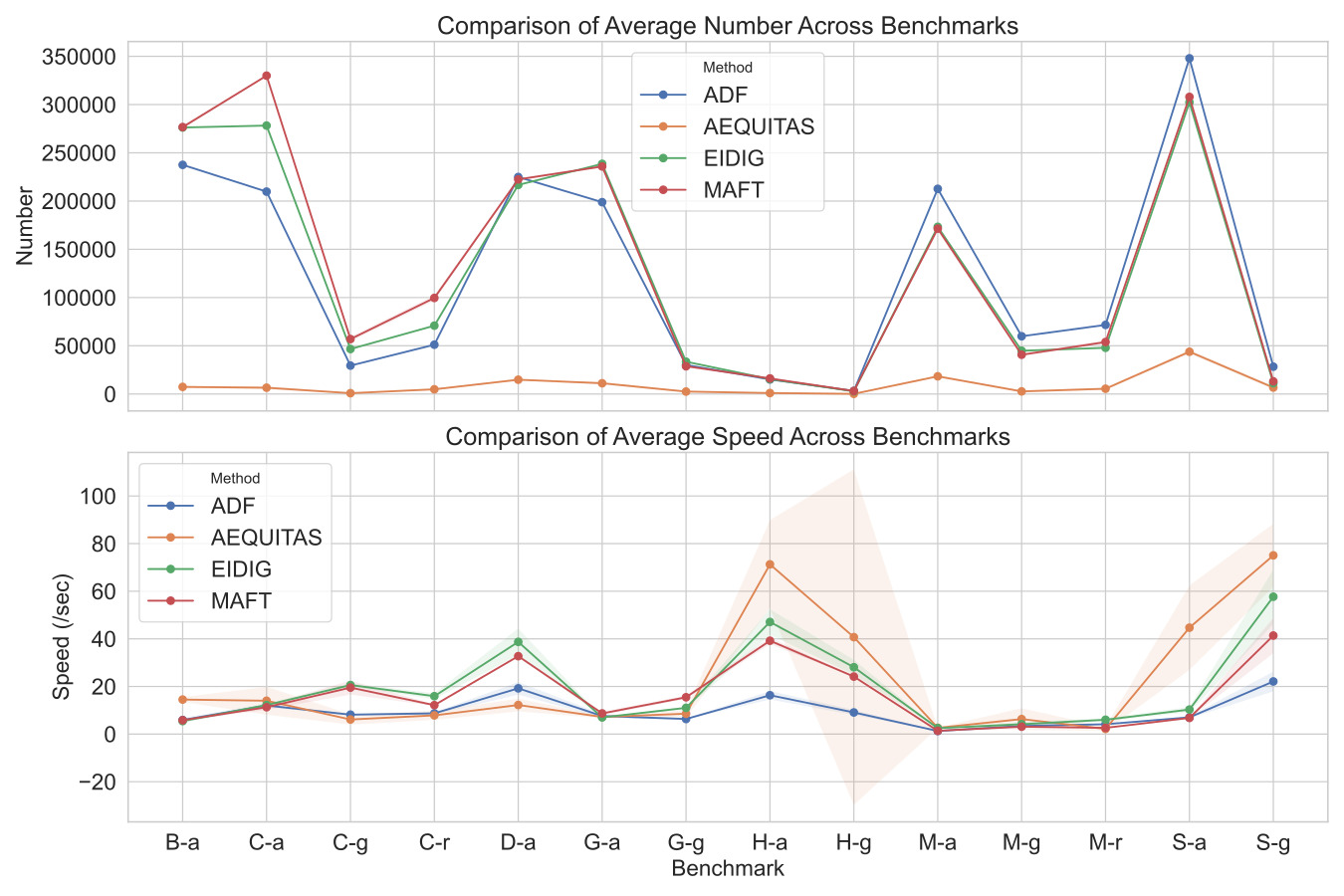}
\caption{\added{Individual Discriminatory Instance Generation Comparison}}
\label{Fig:complete comparision}
\end{figure}



After selecting the hyperparameter, we compare the overall performance of MAFT to the \deleted{benchmarks} \added{baselines} such as \added{AEQUITAS,} ADF and EIDIG, considering both the quantity of generated test cases which means \textbf{effectiveness} and the generation speed which means \textbf{efficiency}. \added{Note that we exclude SG because it is too time-consuming.}

Following the setting established by EIDIG, we select 1000 instance\added{s} from clustered dataset as input\added{s} for global generation, and subsequently generate 1000 instances in the neighbourhood of each global discriminatory seed during local generation (i.e.,$g\_num = l\_num = 1000$). 
Note that \deleted{the German Credit dataset only includes 600 instances,} some datasets are small which number of instances are less than $1000$, so we utilize the whole training set as seeds for global generation. 
\added{For each benchmark, we run three rounds and average the results.}
The comparison results are displayed in Fig \ref{Fig:complete comparision}\added{, where we plot the average results with line charts and mark the fluctuation ranges with shading}. 
The first \deleted{sub figure}\added{subfigure} illustrates the number of individual discriminatory instances generated by each approach, whereas \added{the} second \deleted{sub figure}\added{subfigure} demonstrates the number of individual discriminatory instances generated per second. 
The MAFT \deleted{curve is nearly 100\% identical to} \added{lines are largely consistent with} the EIDIG \deleted{curve} \added{lines}, and both \deleted{consistently outperform the ADF in terms of both the number of discriminations and the speed with which they are generated} \added{achieve comparable effectiveness and much better efficiency than ADF}.
\added{We further apply t-tests \cite{student1908probable} to these methods pairwise. There is no statistically significant difference between the results of effectiveness of ADF, EIDIG, and MAFT, and there is also no statistically significant difference in efficiency of EIDIG and MAFT.}

\deleted{
EIDIG indicates that it generates $19.11\%$ more individual discriminatory instances with a speedup of $121.49\%$ compared to ADF when the number of search attempts is fixed. Meanwhile\added{,} ADF shows that they can generate\deleted{}{s} much more individual discriminatory instances (25 times) using much less time (half to 1/7) than existing black-box methods.
Then we can get that our approach generates more individual discriminatory instances (nearly 30 times) in much less time (6.45\% to 22.57\%) than current black-box approaches.
}
\added{Specifically, MAFT generates 1369.42\% more discriminatory samples with 28.41\% lower speed when compared with AEQUITAS, 7.92\% more discriminatory samples with 70.77\% higher speed when compared with ADF, and 5.58\% more discriminatory samples with 15.91\% lower speed when compared with EIDIG.}
Despite \deleted{removing the fast computation of gradient depends on model itself} \added{avoiding the direct computation of gradient depending on the model}, our approach demonstrates an excellent performance in generating individual discriminatory instances.
\added{As a black-box method, AEQUITAS achieves outstanding performance w.r.t efficiency, but its speed is extremely unstable. Moreover, AEQUITAS generates much less biased samples than the white-box methods, which fails to fully expose the fairness issues of the tested DNNs. Our results are consistent with the results from \cite{zhang2020white}, which prove that AEQUITAS turns inefficient under a target of generating numerous biased samples.}


As to why the estimated gradients achieve similar results to the real gradients in EIDIG in terms of effectiveness and efficiency, We will explore further in the following experiments.

\textbf{In conclusion, MAFT matches the performance of the \deleted{leading} \added{state-of-the-art} white-box method EIDIG \deleted{and} \added{,} significantly outperforms another white-box method ADF\deleted{. Equivalently, MAFT markedly surpass} \added{, and markedly surpasses} other black-box methods like SG and AEQUITAS.}

\subsubsection{Gradient Validation (RQ3)}

\begin{figure*}
\centering 
\begin{tabular}{ccc}
\subfigure[Gradient Cosine similarity.]{
\label{fig:fig1}
\includegraphics[width=0.32\textwidth]{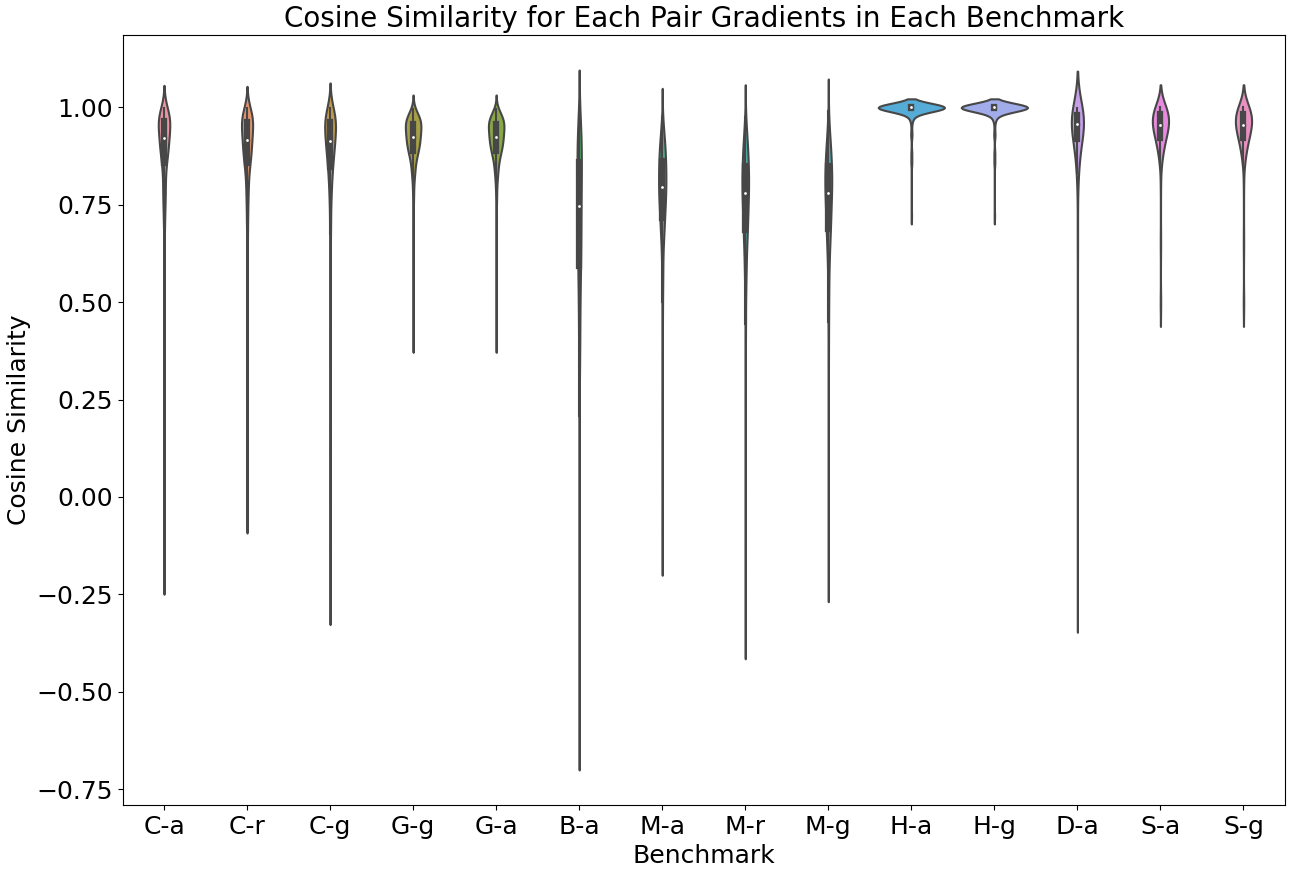}}&
\subfigure[Gradient directions Cosine similarity.]{
\label{fig:fig3}
\includegraphics[width=0.32\textwidth]{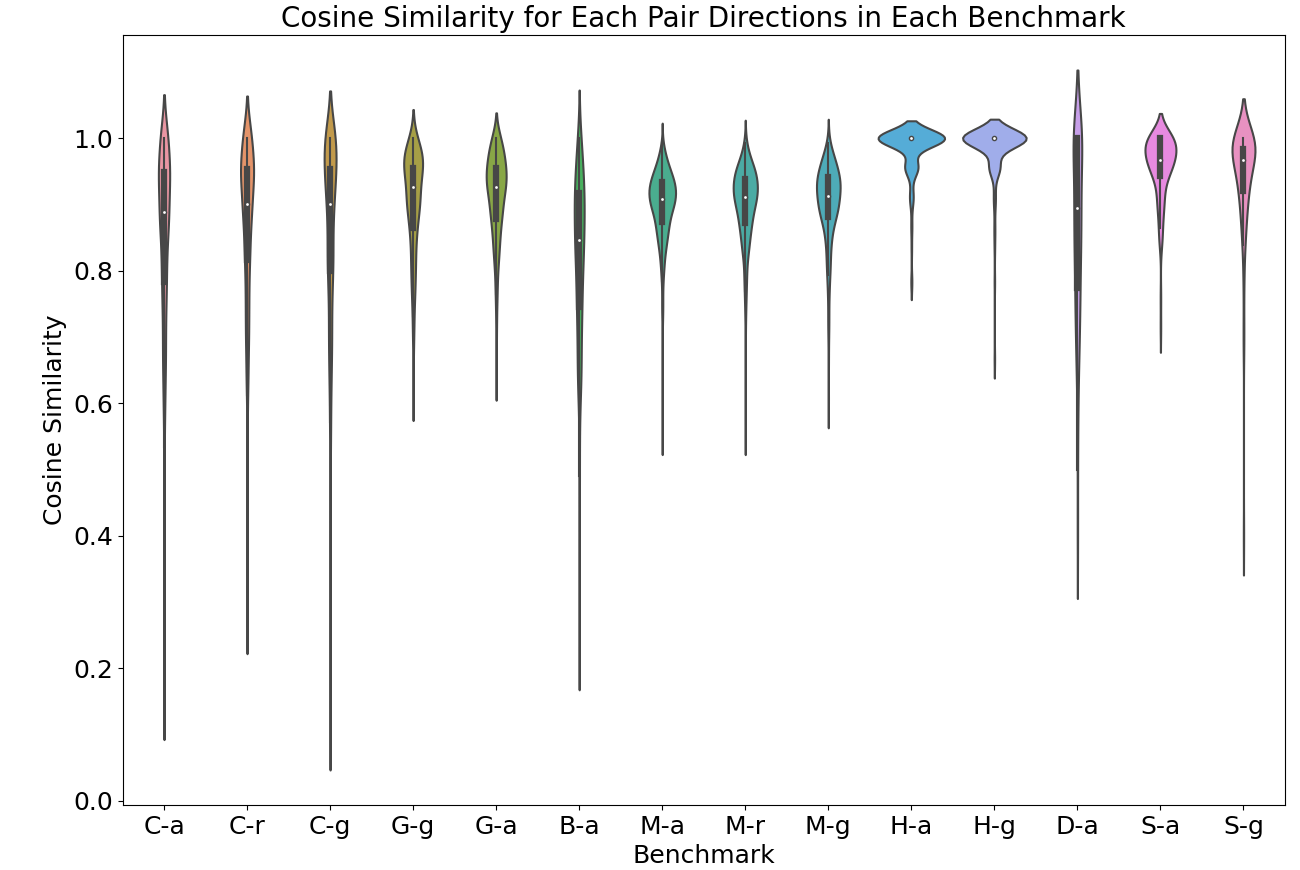}}&
\subfigure[Gradient normalized probability Cosine similarity.]{
\label{fig:fig5}
\includegraphics[width=0.32\textwidth]{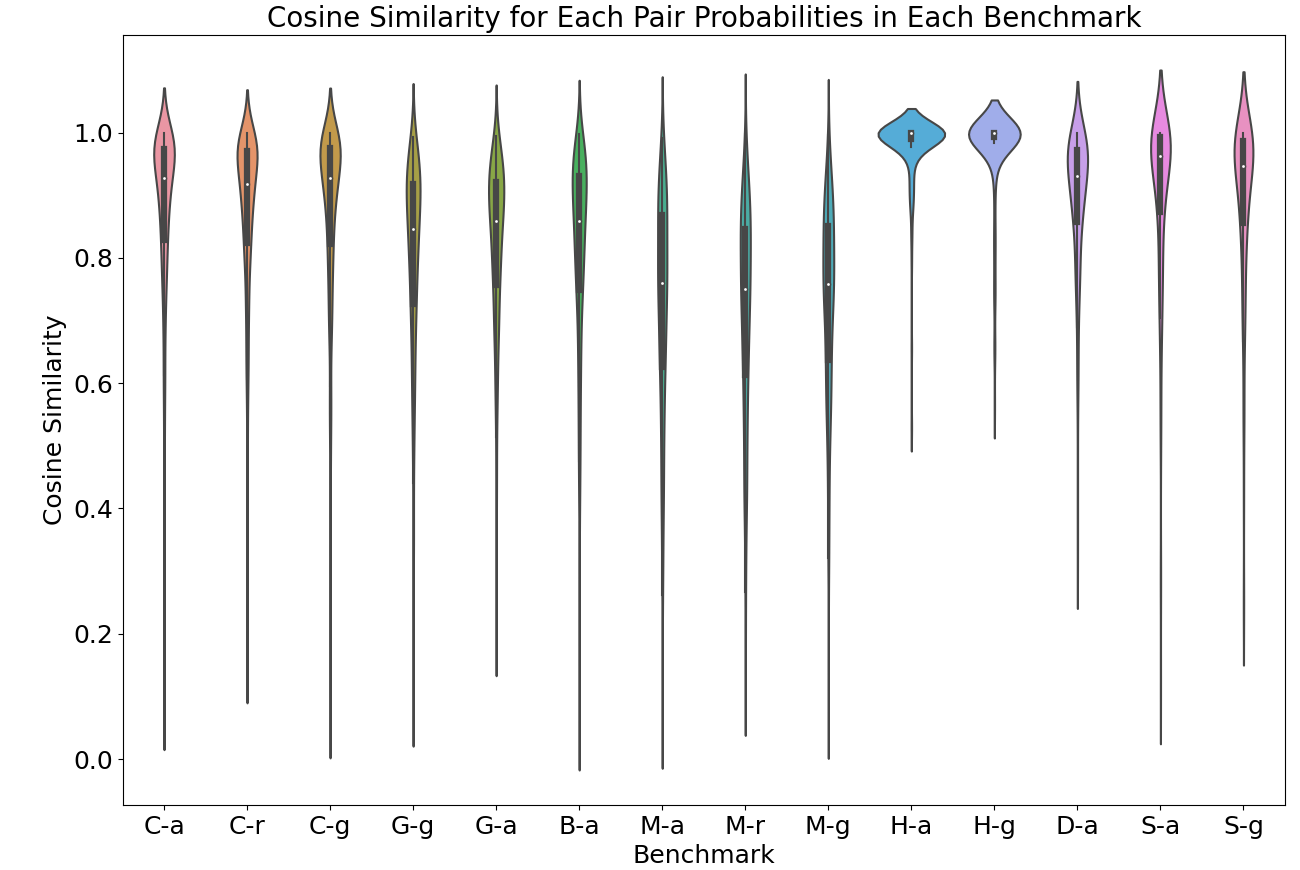}}
\\\vspace{-1mm}
\subfigure[Gradient computation time.]{
\label{fig:fig2}
\includegraphics[width=0.32\textwidth]
{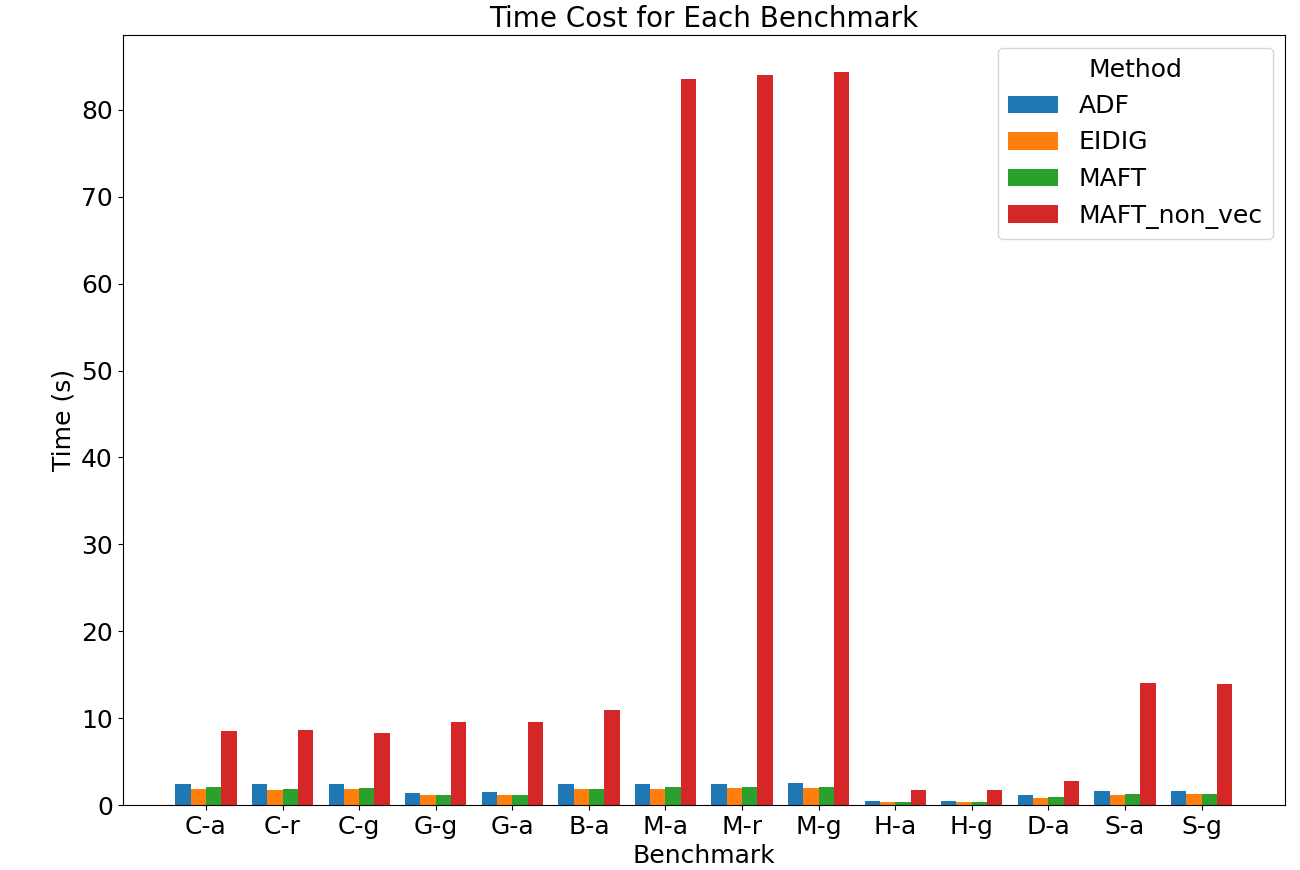}}&
\subfigure[Gradient directions computation time.]{
\label{fig:fig4}
\includegraphics[width=0.32\textwidth]{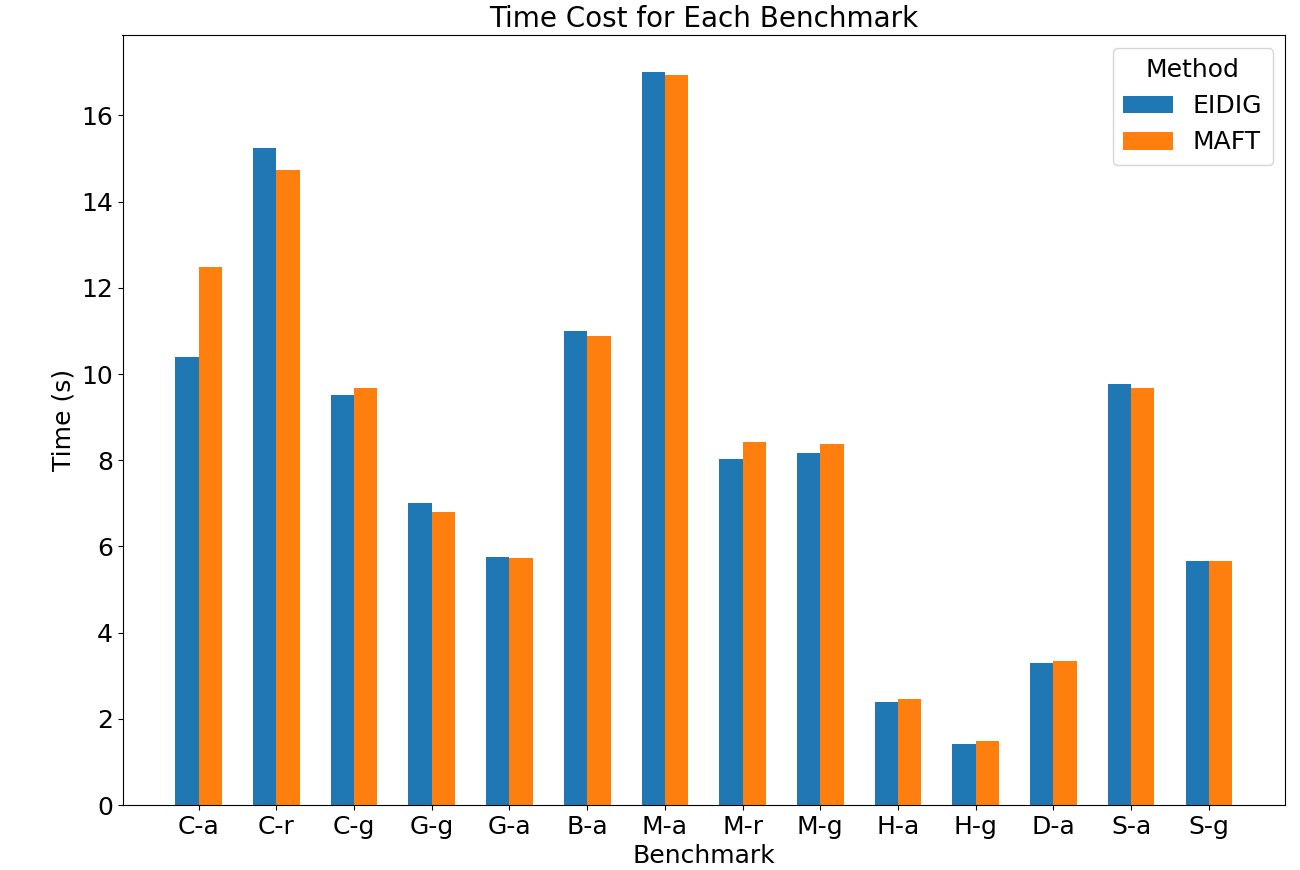}}&
\subfigure[Gradient normalized probability computation time.]{
\label{fig:fig6}
\includegraphics[width=0.32\textwidth]{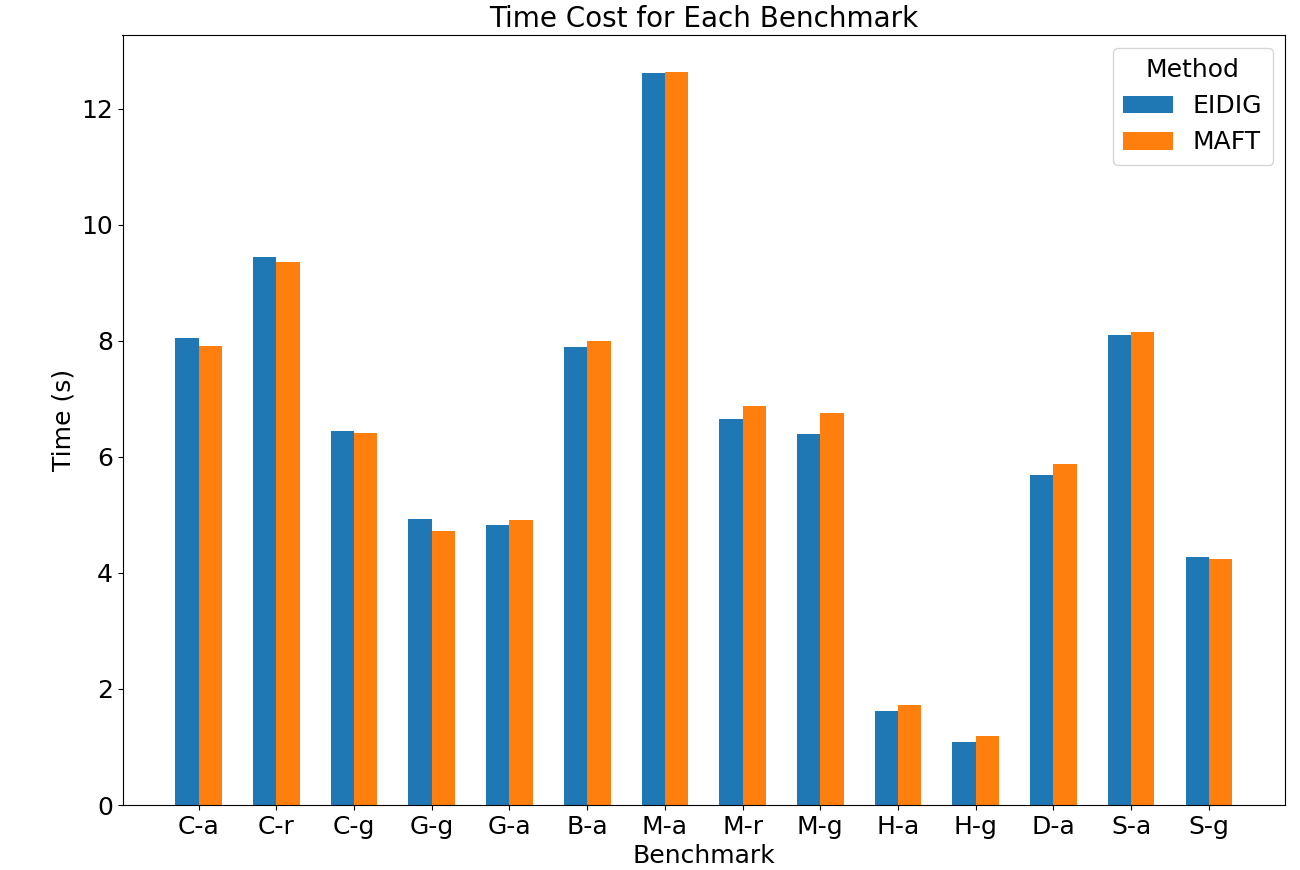}}
\end{tabular}
\label{Fig.gradient comparison}
\caption{\added{Gradient : Comprehensive Comparison}}
\end{figure*}

This question aims to highlight the effectiveness and efficiency of MAFT's gradient estimation procedure by mainly comparing with EIDIG. We not only compare the gradient itself, but also compare the gradient directions and normalized probabilities computed and used by MAFT and EIDIG at both the global and local stages.

The sub-questions in RQ3 are:
\begin{description}
    \item[RQ3.1:] How similar are the estimated gradient of the MAFT and the real gradient of other methods, and how similar are their costs in terms of computational time?
    \item[RQ3.2:] In the global stage, how similar are the guidance directions of the perturbations calculated using the gradient of MAFT and EIDIG, and how similar are their costs in terms of computational time?
    \item[RQ3.3:] In the local stage, how similar are the \deleted{normalization probability}\added{normalized probabilities} of perturbation\deleted{s} calculated using the gradients of MAFT and EIDIG, and how similar are their costs in terms of computational time?
\end{description}

\added{We use cosine similarity as a metric.}
\added{For two vector $\vec{A}$ and $\vec{B}$, the cosine similarity are defined as: 
}
\begin{equation}
    \added{
    \text{cosine similarity}(\vec{A}, \vec{B}) = \frac{\vec{A} \cdot \vec{B}}{\| \vec{A} \| \times \| \vec{B} \|}
    }
\end{equation}
\added{which $\vec{A} \cdot \vec{B}$ is dot product and $\| \vec{A} \| \times \| \vec{B} \|$ is product of vector modules.}
For RQ3.1, we conduct separate experiments to compare the functions for gradient calculation and estimation. For each benchmark, we randomly select 1000 seeds\deleted{(or 600 seeds for the credit dataset)} \added{(or whole small dataset as seeds)} from the original dataset. \deleted{For each seed, the gradient was either computed or estimated. }The similarity between the real gradient and the estimated gradient for each seed \deleted{was} \added{is} then calculated. 
The results are presented in Fig \ref{fig:fig1} \deleted{and Fig \ref{fig:fig2}}.

Regarding effectiveness, it is observed from Fig \ref{fig:fig1} that the average gradient similarity across the majority of benchmarks is above 0.8
and the distribution of them concentrate at the top value (about 1) of the violin graph\deleted{\ref{fig:fig2}}, which means the compactness of their distribution.
This result confirms that the simulated gradient can provide sufficient guidance for exploring the input space and identifying various discriminatory instances. 

\begin{figure}
    \centering
    \includegraphics[scale=0.2]{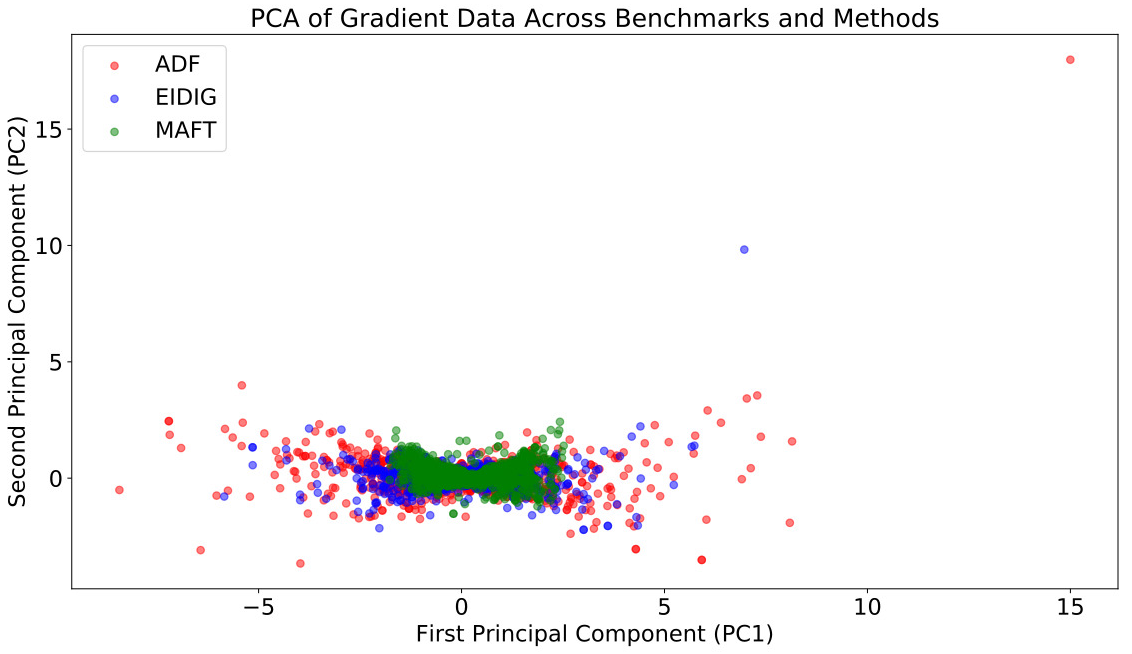}
    \caption{\added{Gradient PCA Comparision}}
    \label{Fig:grandient_pca}
\end{figure}

\added{We also process the gradients to project them onto a two-dimensional space through Principal Component Analysis (PCA) \cite{mackiewicz1993principal} and illustrate the intuitive result in Fig \ref{Fig:grandient_pca} which axes represent the pricipal componenents of PCA. This visualization effectively demonstrates the close alignment of gradients produced by MAFT with those from EIDIG, and a distinct deviation from the gradients generated by ADF due to the transformation from loss gradients to output probability gradients.}

Regarding efficiency, Fig \ref{fig:fig2} displays the time consumption for ADF, EIDIG, MAFT and MAFT\_non\_vec (MAFT\_non\_vec uses original estimation algorithm shown as Algorithm \ref{alg:zoo_non_vec}). 
\added{MAFT\_non\_vec costs excessive amount of time on MEPS15, because MEPS15 has over 100 attributes, which are much more than those of others.}
EIDIG is noticeably faster than ADF which can be attributed to the absence of the backpropagation through the loss function\deleted{ and thus it perform total experiment faster than the latter}. 
\deleted{As for why}\added{Why} MAFT takes slightly more time than EIDIG, but the performance is almost the same?
The calculation of 1000 gradients \added{costs} less than 10 seconds (except MAFT\_non\_vec), which is a relatively minor part of the entire exploration\deleted{ scale}. This will be shown \deleted{more obvious}\added{obviously} in next two comparison\added{s}.
The time cost difference between MAFT and EIDIG is negligible\deleted{ especially with additional computational operations when} consider\added{ing} the entire exploration process. 
\textit{Time \deleted{Efficiency}\added{Complexity} Analysis of Gradient Estimation}.
Let's recall the \deleted{time efficiency}\added{process} of getting real gradient by backpropagation (BP): First, backpropagation requires once forward propagation (FP) to compute the output $f(x)$. Second, it requires once backward propagation \deleted{pass} to compute the gradients of the loss (ADF) or output value (EIDIG) w.r.t \added{a} specific input. So the complexity for this step is O(FP + BP). 
While in estimation algorithm, we can forward \deleted{propagation}\added{propagate} twice on the input $X$ and $x$ separately through the model to compute zero-order gradient. The time complexity for this step is just O(2*FP) ignoring fixed computation operation\added{s}. 
Subtracting the common forward propagation step of those two methods, we can focus on remaining computational effort\added{s} with the vectored zero-order gradient method at O(FP) and backpropagation at O(BP). 
It's \deleted{easy to know}\added{obvious} that forward propagation should just through the model \deleted{form}\added{of} the input layer to output layer in the forward direction. According chain rule, backpropagation computes gradients directly for each layer, requiring only a single pass through the model in the opposite direction from forward propagation.
The relationship between O(FP) and O(BP) depends on the specific model and implementation, but typically, O(FP) and O(BP) are of similar orders of magnitude since both methods \deleted{involving}\added{involve} traversing \deleted{through} the layers of model once.


Next, regarding RQ3.2 and RQ3.3, further exploration is conducted on the directions and \deleted{normalization}\added{normalized} \deleted{probability}\added{probabilities} of perturbation that serve to guide exploration during the global and local stages, respectively.

In the case of RQ3.2, we \deleted{modify the global generation method to} solely compare the consistency of direction information used in \added{the} first iteration of global generation. For each benchmark, we randomly select 1000 seeds\deleted{(600 for the credit dataset)} \added{(or whole small dataset as seeds)}. 
For each instance \deleted{$x1$}\added{$x_1$} from the seeds, we identify its most dissimilar instance \deleted{$x2$}\added{$x_2$} \deleted{for it}\added{from the similar set which contains  all instances that differ only in protected attributes from $x_1$}, to construct a pair \deleted{$(x1, x2)$}\added{$(x_1,x_2)$}. 
\deleted{If they are not a pair of discriminatory instances, we calculate their real gradients and estimated gradients and obtain two sets of perturbation directions by comparing the signs of these gradients. Otherwise they will be ignored.
The \deleted{similarities}\added{similarity} between these real directions and estimated directions is further computed and compared.}
\added{If they are a pair of discriminatory instances, they will be ignored as there is no need for further global search.
Otherwise, we calculate their real gradients denoted as $(g_1, g_2)$ and estimated gradients denoted as $(\Tilde{g_1}, \Tilde{g_2})$. 
Subsequently, we obtain the real gradient direction $d$ by comparing the signs of $g_1$ and $g_2$, and estimated gradient direction $\Tilde{d}$ by comparing $\Tilde{g_1}$ and $\Tilde{g_2}$.
The iteration continues until completion for each seed, 
resulting two sets of real directions and their corresponding estimated directions, respectively.}
The \deleted{similarities}\added{similarity} between these real directions and estimated directions is further computed and compared.

As for RQ3.3, we \deleted{also modify the local generation method to} compare the consistency of \deleted{normalization}\added{normalized} probability information used in \deleted{global}\added{local} generation.
We use randomly collected 1000 seeds instead of globally discriminatory instances as inputs and construct pair\added{s} for them like \deleted{in} RQ3.2. 
\added{Subsequently, we independently}
\deleted{We then} calculate the real and estimated gradients for the pair\added{ed} instances.
\added{Following this, we}\deleted{and} compute the \deleted{normalization probability pair be}\added{normalized probabilities} \added{denoted as $p$ and $\Tilde{p}$, used in selecting the perturbation attributes}.
\added{The similarity of them is also be computed and compared.}

We find that \deleted{the perturbation directions similarity}\added{the similarity of perturbation directions} and \deleted{normalization probability}\added{the similarity of normalized probabilities} are highly consistent with \deleted{gradients similarity}\added{the similarity of gradients,} which means the fine-grain\added{ed} usage of estimated gradient is still useful.
\deleted{Average gradient directions cosine similarity and directions distribution}\added{The cosine similarity distribution of gradient directions} in \deleted{\ref{fig:fig4} and \ref{fig:fig5} are} \added{Fig \ref{fig:fig3} is} consistent with that in \deleted{\ref{fig:fig1} and \ref{fig:fig2}} \added{Fig \ref{fig:fig1}}\deleted{separately. Average normalization probability similarity and distribution in \ref{fig:fig7} and \ref{fig:fig8} are also the same as \ref{fig:fig1} and \ref{fig:fig2} separately.}\added{, so is it in Fig \ref{fig:fig5}.}

The time comparison in \deleted{\ref{fig:fig6} and \ref{fig:fig9}} \added{Fig \ref{fig:fig4} and Fig \ref{fig:fig6}} demonstrate that when considering both gradient calculation and other computational operations within a single iteration, the time difference between EIDIG and MAFT further decreases.


\textbf{In summary, the experimental results from RQ3.1 to RQ3.3 convincingly demonstrate that \added{the} estimated gradient\deleted{ in terms of both efficiency and effectiveness} closely matches the real gradient \added{in terms of both efficiency and effectiveness}. }

\subsection{Threat to Validity}

\paragraph{\added{Internal Validity}}
\added{We take average of repeated experiments to ensure the validity of our conclusion. We consider only one sensitive attribute as each benchmark for our experiments. Actually, we have experimentally verified that MAFT remains superior to the other methods considering multiple sensitive attributes. All methods relatively generate more biased samples with more execution time in such circumstances, since all the possible combinations of their unique values need to be checked.}

\paragraph{\added{External Validity}}
\added{Our black-box method is naturally applicable to more scenarios than white-box methods, and the gradient estimation technique of MAFT could be combined with follow-up testing frameworks based on gradient. To ensure the generalization of our results, we conduct experiments on 7 public tabular datasets frequently used in fairness research. For unstructural datasets, such as images or texts, the key aspect is to design reasonable attribute flip methods before applications of MAFT. We refer interested readers to related works adapted for images\cite{zhang2021fairness,zheng2022neuronfair,xiao2023latent} and texts\cite{ma2020metamorphic,soremekun2022astraea,fan2022explanation}, which could be integrated with our method. In addition, we have only tested fully-connected networks, since the shallow fully-connected networks are good enough to accomplish the subject prediction tasks. Theoretically, MAFT is applicable for testing of more complicated models, like CNNs or RNNs, if the tested models are differentiable or differentiable almost everywhere. We will extend MAFT to unstructural datasets with diverse model architectures in future. For the strict settings where predicted probabilities are not available, a simple but coarse solution is to substitute the probability vector with a one-hot vector, of which the element corresponding to the predicted label is set to 1, and more sophisticated tricks need to be explored in future.}

\vspace{-2mm}
\section{Related Work}
\label{sec:related work}
\paragraph{Fair Deep Learning}
Recently, many literatures have \deleted{identify}\added{identified} \deleted{and divide} several causes that can lead to unfairness in Deep Learning\deleted{ into two categories:}\added{, including} data and model. \deleted{At the same time, many}\added{Many} papers have proposed mechanisms to improve the fairness of ML algorithms. These mechanisms are typically categorised into three types \added{\cite{pessach2022review}}: pre-process\added{ing}\added{\cite{calmon2017optimized, font2019equalizing, louizos2015variational, samadi2018price, zemel2013learning}}, in-process\added{ing}\added{\cite{agarwal2018reductions, bechavod2017learning, bechavod2017penalizing, calders2010three, goh2016satisfying}}, and post-process\added{ing}\added{\cite{corbett2017algorithmic, dwork2018decoupled, hardt2016advances, may2019measuring}}. \deleted{\cite{pessach2022review}}
\deleted{Pre-process\added{ing} machanisms \cite{calmon2017optimized, font2019equalizing, louizos2015variational, samadi2018price, zemel2013learning} involve changing the training data to remove bias from them before training. Many mechanisms suggest\deleted{s} that fair feature representations and fair data distribution make fair classifier.  In-process\added{ing} methods \cite{agarwal2018reductions, bechavod2017learning, bechavod2017penalizing, calders2010three, goh2016satisfying} try to modify the learning algorithm to get fair model during the training time, for example\added{,} \deleted{regularize}\added{regularizing} the objective function. Post-process\added{ing} methods \cite{corbett2017algorithmic, dwork2018decoupled, hardt2016advances, may2019measuring} perform calibration on the output scores to make decisions fairer by enforcing them satisfying specific fairness metrics.}
\added{There are also some fairness research focusing on specific domains, such as news recommendation\cite{wu2021fairness} and recruitment\cite{emelianov2020fair, emelianov2022fair}.}

\deleted{
On the other hand, there are authors who focus on studying domain-specific fairness approaches. For example, as the development of online news services, Wu et al. \cite{wu2021fairness} proposes a fairness-aware news recommendation approach with decomposed adversarial learning and orthogonality regularization.
Vitalii et al. \cite{emelianov2020fair, emelianov2022fair}, focus on the discrimination in selection processes such as recruitment or university admissions and show that imposing the $\gamma\_rule$(it is an fairness rule) increases the selection utility when the noise variances of different gourps of candidates are unknown to the decision makers while slightly decreases the selection utility when the noise variances are known(which is ideal optimal settings but not realistic).
}

\vspace{-2mm}

\paragraph{Fairness Testing}
\sloppy
Bias in software is a prevalent issue, even when fairness is explicitly considered during the design process. Therefore, conducting fairness testing is a crucial step \deleted{that should be undertaken} before deployment\cite{galhotra2017fairness, huchard2018proceedings}.
Some works highly relative to this work have been introduced in Section \ref{sec:background}.
In addition to those, there are also some other works.
Zheng et al. \cite{zheng2022neuronfair} proposed a white-box fairness tesing method Neuron\deleted{-}Fair which can handle both structured and unstructured data and \deleted{can} quantitatively interpret\deleted{s} DNN\deleted{s'}s fairness violations. 
Ma et.al \cite{ma2022enhanced} introduce\deleted{s}\added{d} a seed selection approach I\&D focusing on generating effective initial individual discriminatory instances to enhance fairness testing.

\vspace{-1mm}
\section{Conclusion}
\label{sec:conclusion}
We propos a novel black-box individual fairness testing method called Model-Agnostic Fairness Testing (MAFT). MAFT allows practitioners to effectively identify and address discrimination in DL models, regardless of the specific algorithm or architecture employed. \deleted{Our method provides a more flexible approach to deep learning fairness testing by eliminating the need for backpropagation and other model-specific techniques. } \deleted{Our approach combines a two-stage generation framework to effectively search and identify discriminatory individual instances under the guidance of the estimated gradient.
}
\deleted{We extensively evaluated the performance of MAFT on various benchmark\added{s}.} 
The experimental results demonstrate that MAFT \deleted{could even} achieve\added{s} the same effectiveness as state-of-the-art white-box methods whilst \deleted{significantly} \deleted{reducing the time cost and} improving the applicability to large-scale networks.  
Compared to existing black-box approaches, \deleted{MAFT  generated up to 30 times more individual discriminatory instances with only 6.45\% - 22.57\% time cost. Thanks to the black-box feature of our approach, we believe MAFT is also applicable to other AI models besides DNNs if the training data satisfies the assumption of our approach.}
\added{our approach demonstrates distinguished performance in discovering fairness violations w.r.t effectiveness ($\sim14.69\times$) and efficiency ($\sim32.58\times$).}



\section*{Acknowledgments}
\sloppy
This work is jointly supported by the NSFC-ISF Project 
(No. 62161146001) and the 'Digital Silk Road' Shanghai International Joint Lab of Trustworthy Intelligent Software (No. 22510750100), with additional funding from the Shanghai Trusted Industry Internet Software Collaborative Innovation Center.
Additionally, we express our sincere thanks to Lingfeng Zhang, the first author of the EIDIG work, for his invaluable assistance in refining our experiments and improving academic manuscript during the revision process.

\clearpage 
\balance
\bibliographystyle{unsrt}  
\bibliography{references}

\end{document}